\documentclass[10pt,twocolumn,letterpaper]{article}

\usepackage{cvpr}      

\usepackage{bm}
\usepackage{amsfonts}
\usepackage{graphicx}
\usepackage{amsmath}
\usepackage{amssymb}
\usepackage{booktabs}
\usepackage{diagbox}
\usepackage{setspace}
\usepackage{xspace}
\usepackage{multirow}
\usepackage{algorithm}
\usepackage{algpseudocodex}
\usepackage{cuted}
\usepackage{fontawesome}
\usepackage{tikz}
\usepackage{lipsum} 
\newcolumntype{M}[1]{>{\centering\arraybackslash}m{#1}}
\usepackage{soul}
\usepackage[pagebackref,breaklinks,colorlinks,allcolors=cvprblue]{hyperref}

\hypersetup{
    urlcolor=cvprpink 
}
\begin{document}



\definecolor{cvprblue}{rgb}{0.21,0.49,0.74}
\definecolor{cvprpink}{rgb}{1,0.3,0.75}

\title{Acc3D: Accelerating Single Image to 3D Diffusion Models via Edge Consistency Guided Score Distillation}

\author{
    Kendong Liu\textsuperscript{1}  \quad\quad
    Zhiyu Zhu\textsuperscript{1}\thanks{Corresponding author and Equation contribution.}  \quad\quad
    Hui Liu\textsuperscript{2}  \quad\quad
    Junhui Hou\textsuperscript{1}\thanks{This work was supported in part by the NSFC Excellent Young Scientists Fund 62422118, in part by the Hong Kong
    RGC under Grants 11219324 and 11219422, and in part by the Hong Kong UGC under Grants UGC/FDS11/E02/22 and UGC/FDS11/E03/24.} \\
    \textsuperscript{1}City University of Hong Kong~\quad
    \textsuperscript{2}Saint Francis University \\
    \texttt{\{kdliu2-c,zhiyuzhu2-c\}@my.cityu.edu.hk} \\
    \texttt{h2liu@sfu.edu.hk}~\texttt{jh.hou@cityu.edu.hk} \\
    \url{https://acc3d-object.github.io/}
}

\maketitle

\begin{strip}
    \vspace{-8em}
    \centering
    \includegraphics[width=1.0\textwidth]{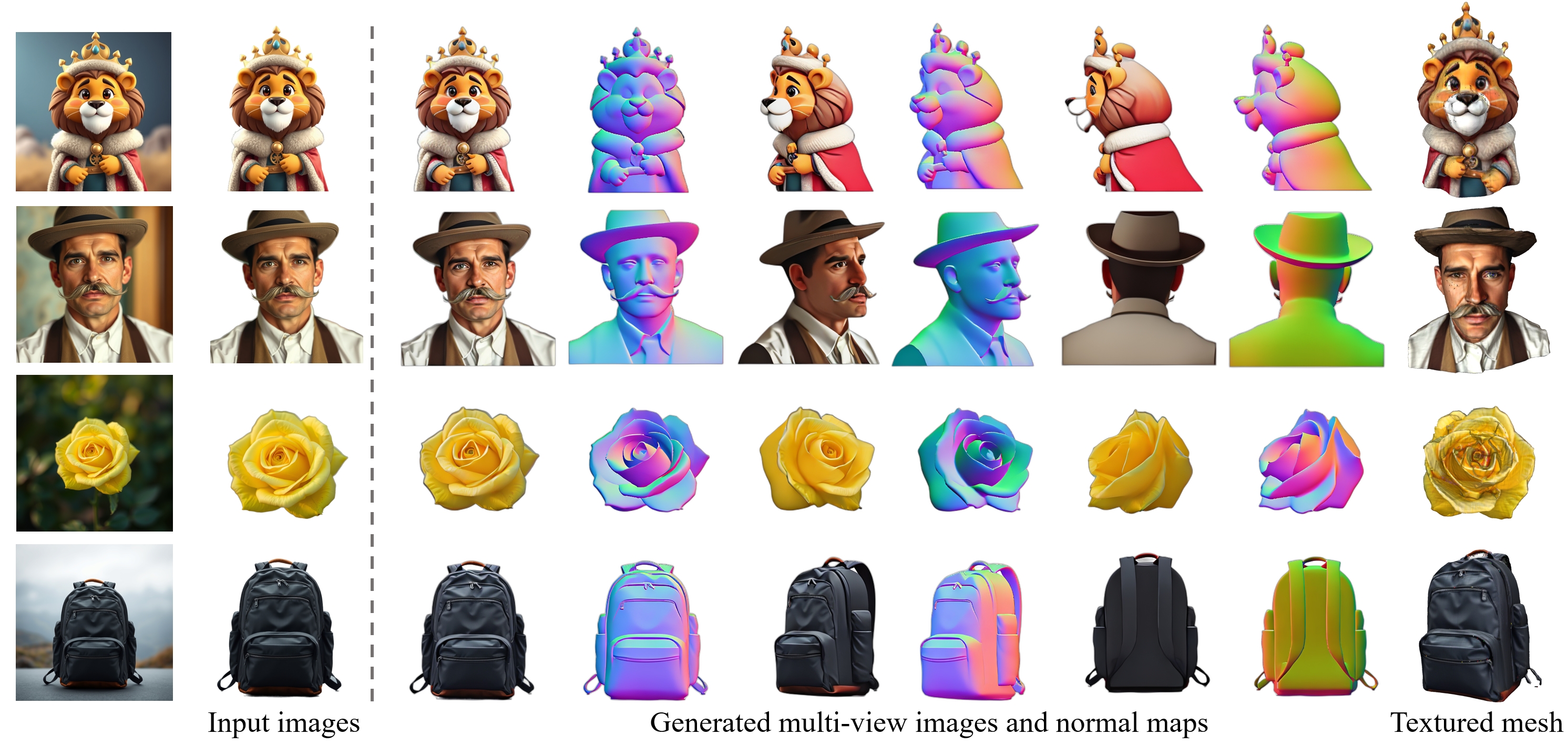}
    \vspace{-2em}
    \captionof{figure}{Visual illustration of the generated high-quality multiview images and normal maps from a given single-view image by our Acc3D through \textbf{fewer than four} inference steps. \color{blue}{\faSearch~} Zoom in for details. 
    }
    \label{fig:teaser} \vspace{-1em}
\end{strip}

\begin{abstract}
We present Acc3D to tackle the challenge of accelerating the diffusion process to generate 3D models from single images. To derive high-quality reconstructions through few-step inferences, we emphasize the critical issue of regularizing the learning of score function in states of random noise. To this end, we propose edge consistency, i.e., consistent predictions across the high signal-to-noise ratio region, to enhance a pre-trained diffusion model, enabling a distillation-based refinement of the endpoint score function. Building on those distilled diffusion models, we propose an adversarial augmentation strategy to further enrich the generation detail and boost overall generation quality. The two modules complement each other, mutually reinforcing to elevate generative performance. Extensive experiments demonstrate that our Acc3D not only achieves over a $20\times$ increase in computational efficiency but also yields notable quality improvements, compared to the state-of-the-arts. 
\end{abstract}

\vspace{-1.5em}
\section{Introduction}
\vspace{-1mm}
\label{sec:intro}

Single image-based 3D reconstruction stands as a pivotal domain within the realms of 3D computer vision and computer graphics~\citep{dou2023tore, magic123, point-e, syncdreamer}, boasting extensive applications in virtual reality, 3D gaming, content creation, and precision robotics. Despite humans' innate ability to perceive three-dimensional structures from a single image, the swift and accurate generation of consistent content remains a formidable challenge.

Diffusion models~\citep{ho2020denoising} have demonstrated their strong capability in image generation and video synthesis~\cite{liu2024prefpaint}, paving the way for diffusion-based 3D content creation. Various studies~\citep{DreamField, DreamFusion, magic123} have been dedicated to distilling consistent 3D representations, such as neural radiance fields (NeRF)~\citep{nerf} or 3D Gaussian splatting~\cite{gaussiansplatting}, from 2D image diffusion models or vision language models using the Score Distillation Sampling (SDS) loss~\citep{DreamFusion}.
Although these methods yield visually pleasing results, the distillation process tends to be time-intensive for generating a single shape, requires intricate parameter tuning to obtain satisfactory quality, and often faces issues with unstable convergence and quality degradation. Another stream of research~\citep{pointclouds, mesh} aims to directly generate 3D geometries, such as point clouds and meshes. However, the necessity for large-scale, open-source, native 3D samples inevitably hampers the development of these methods.

Recent research, exemplified by Syncdreamer~\cite{syncdreamer}, Wonder3D~\cite{wonder3d}, MVDream~\cite{mvdream}, and Era3D~\cite{era3d}, has focused on generating multiview-consistent images directly to facilitate single-view 3D reconstruction of diverse objects. These approaches leverage the pre-trained diffusion pipeline for single image generation to model the joint probability distribution of multiview-consistent images. By improving the multiview consistency in image generation, these methods can reconstruct 3D shapes from the produced multiview images using neural reconstruction methods~\citep{wang2021neus}. To improve the accuracy of 3D object reconstruction, the diffusion frameworks used in Wonder3D and Era3D directly generate normal maps and multiview images. These are then exploited in tandem with both normal maps and multiview images to reconstruct the 3D object. However, the lengthy sampling time required for the integration of the reversed differential equation burdens all these diffusion-based multiview image generation methods. This typically involves an iterative process that progressively denoises a Gaussian noise sample into an image, emphasizing the importance of \textit{sampling acceleration}.

In this paper, we tackle this challenge by introducing a consistency-based endpoint score-matching mechanism to achieve the few-step generation, which focuses on enhancing the accuracy of score function in low-SNR regions. 
Specifically, to regularize the score function for greater accuracy, we employ consistency training techniques to enforce stable score estimation in high-SNR regions, which can act as a score corrector, refining the coarse generation from the endpoint pure noise state.
We subsequently introduce an adversarial training technique to further boost the alignment of the manifold of generated samples~\cite{GANs}, where the pre-trained diffusion model is also introduced as a discriminator to fully leverage the pre-trained geometric knowledge. Our Acc3D is composed of two main components: \textit{edge consistency-guided distillation} and \textit{disentangled adversarial regularization}. 
These two components complement each other—distillation stabilizes adversarial training, reducing the risk of mode collapse, while adversarial learning enhances the perceptual richness of the model. Working in tandem, these components create a balanced and sophisticated model that delivers both stability and detail.

Comprehensive experiments show that our accelerated generative model outperforms the baseline model and most other image-to-3D diffusion models. It offers high-quality 3D content across multiple views and provides faster inference. Remarkably, even with only a few iterative steps, our accelerated model can generate high-quality multiview images from a variety of 2D images with different styles.

In summary, the main contributions of this work are as follows.
\begin{itemize}
\item we have proposed an edge consistency-based paradigm to achieve distillation-based acceleration of single image-to-3D diffusion models, driving generation speeds up to $20 \times$ while simultaneously elevating performance; 

\item we have conducted a comprehensive analysis, incorporating both intuitive and theoretical manners to ensure the technical soundness of the proposed method; and

\item we have carried out extensive experiments to demonstrate the performance of the proposed algorithm on both synthetic and real datasets.
\end{itemize}

\begin{figure*}[t]
  \centering
   \includegraphics[width=0.9\linewidth]{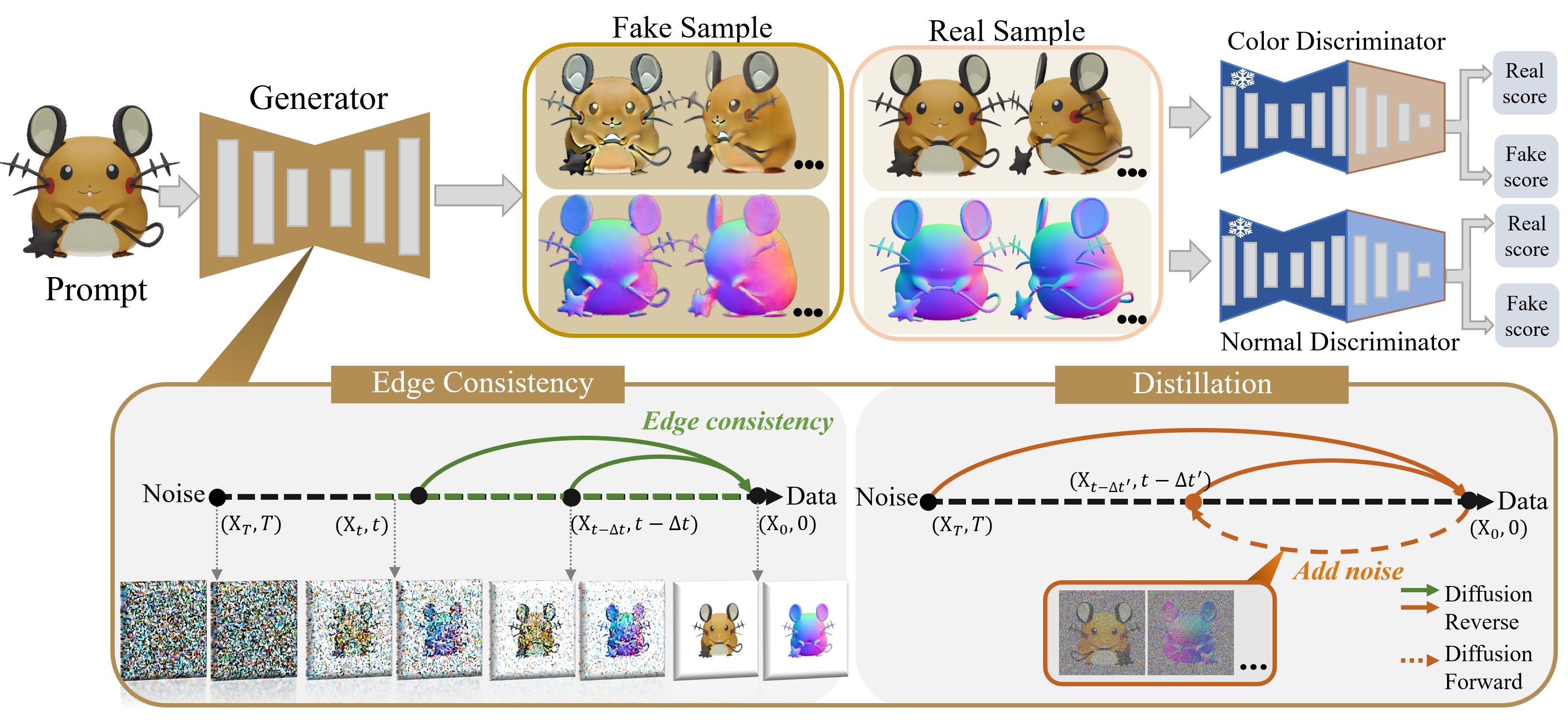}\vspace{-0.2cm}
   \caption{Overview of our Acc3D. The training pipeline unfolds in two core components: \textit{edge consistency-guided distillation} and \textit{adversarial training}. Each component bolsters the other's advantages—the distillation procedure stabilizes adversarial training, mitigating the risk of mode collapse, while adversarial learning can enhance perceptual richness. Collectively, these elements craft a balanced, refined model that excels in both stability and detail.}
   \label{fig:pipeline} \vspace{-0.5cm}
\end{figure*}

\section{Related Work}

\label{sec:related_works}
\noindent\textbf{Image-to-3D Generative Models.}
Diffusion models~\citep{ho2020denoising} have achieved substantial success in 2D image generation, prompting various efforts to extend pre-trained 2D diffusion models to 3D generation. DreamFusion~\cite{DreamFusion} marks a pioneering endeavor in distilling a 2D image generation model to craft 3D content from a single text prompt. Building upon this groundwork, Magic3d~\cite{makeit3d} generates 3D shapes from single images, refining the 3D representation for each instance.  These methods~\citep{sds3, sds2, sds4, magic123} commonly employ the SDS loss to guide the optimization of their 3D representations, such as NeRF, mesh or Gaussian Splatting. Notably, NVS-Solver~\citep{mengnvs} utilizes video diffusion models integrated with data manifold constraints to directly generate consistent multi-views for both static and dynamic scenes.

In order to produce 3D content without the necessity of training each sample, numerous existing studies~\cite{3daware, multiview1, wonder3d, zero123, syncdreamer, mvdream, era3d, zero123++} directly generate multiview images from a single-view image. These image-to-3D diffusion models, trained on a large-scale 3D object dataset~\cite{objaverse}, exhibit impressive generalization capabilities.
Syncdreamer~\cite{syncdreamer} synchronizes the intermediate states of multiview images at every step of the reverse process through a 3D-aware fusion. MVDream~\cite{mvdream} introduces self-attention into multiview images in order to improve consistency. Wonder3D~\cite{wonder3d} generates both multiview normal maps and the corresponding images under a cross-domain self-attention mechanism. Following Wonder3D, Era3D~\cite{era3d} implements a camera prediction module to alleviate image shape distortions and increases the image resolution up to 512$\times$512. 
While these image-to-3D models typically demand a long inference time, our accelerated generative model stands in contrast. It can generate high-quality multiview images and normal maps with fewer iterative processes.

\noindent\textbf{Diffusion Model Acceleration.}
The iterative process of progressive denoising is a key characteristic of diffusion models, making the acceleration of the sampling process a significant area of interest~\cite{luparasolver}. Several studies~\citep{consistency1, consistency2, phased_consistency, consistency3, consistency4} have focused on direct noise-to-data mapping with the objective of generating high-quality images in a single step. The study by \citet{consistency1} introduces a consistency model that condenses the diffusion model process to a single step, training a one-step model to emulate the multi-step outcomes of the original model.
\citet{consistency4}, on the other hand, aims to keep the image generation trajectory as linear as possible between the final and current points.

\noindent\textbf{Generative Adversarial Networks.} 
Several recent studies~\cite{diffusiongan, consistency_trajectory, ladd, add, ufogen, dmd2} have adopted a generative adversarial training approach to maintain and boost the outcomes of accelerated diffusion models. Notably, ~\cite{diffusiongan} incorporates feedback from the discriminator by backpropagating it through the forward diffusion chain.
The length of this chain is adaptively adjusted to strike a balance between noise and data levels. ~\citet{consistency_trajectory} combined GANs and denoising score matching loss to improve overall performance.
~\citet{add, ladd} used the pre-trained diffusion model as the discriminator and updated the parameters in both raw and latent space.
~\citet{dmd2} designed the discriminator similar to~\citep{add}, minimizing the difference between the generator and pre-trained model by computing the KL divergence of two score functions. 
Despite the abundance of methods aimed at achieving stable adversarial training, GANs are still prone to issues such as model collapse and unstable training, which makes scaling and adapting to different distributions challenging.

\section{Preliminary}

In this section, we introduce the basic knowledge of stochastic differential equation (SDE)-based formulation of diffusion models, its integrator, and consistency models.

\vspace{0.4em}
\noindent \textbf{Diffusion Models.} Denote by $\mathbf{X}_t$ the latent noised variable, $t \in [0, T]$ as the scalar indicating the time-stamp. Then, we can formulate the forward diffusion process as the following SDE, gradually shifting the clean image $\mathbf{X}_0$ towards a random noise $\mathbf{X}_T$:\vspace{-0.5em}  
\begin{equation}
\label{eq:SDE}
    d\mathbf{X}_t = f(t)\mathbf{X}_t dt + g(t)d \bm{\omega},\vspace{-0.5em} 
\end{equation}
where $d\bm{\omega}$ represents the standard Wiener process, with $f(t)$ signifying the drift coefficient function, and $g(t)$ denoting the diffusion coefficient function. By reversing the forward SDE given in Eq.~\eqref{eq:SDE}, we can generate the corresponding clean latent from the easily sampled random noise. This process results in the subsequent ordinary differential equation (ODE) formulations:\vspace{-0.5em} 
\begin{equation}
\label{eq:RevODE}
d\mathbf{X}_t = [f(t)\mathbf{X}_t   - \frac{1}{2} g^2(t)\nabla_{\mathbf{X}_t} \log   p(\mathbf{X}_t  ) ] dt,\vspace{-0.5em} 
\end{equation}
where  $\nabla_{\mathbf{X}_t} \log p(\mathbf{X}_t)$ denote the data gradient, usually approximated by a learnable score function of $\mathbf{S}_{\bm{\theta}}(\mathbf{X}_t)$ parameterized with $\bm{\theta}$. 
Consequently, the analytical solution of Eq.~\eqref{eq:RevODE} to the arbitrary timestamp $t$ can be expressed as\vspace{-0.5em}  
\begin{equation}
\label{eq:Solution}
\mathbf{X}_t = \mathbf{X}_T + \int_T^t \frac{d \mathbf{X}_t}{dt} dt,  X_T \sim \mathcal{N} ( \mathbf{0} ,  I),\vspace{-0.5em} 
\end{equation}
where $\mathcal{N} ( \mathbf{0} ,  I)$ denotes  the standard Gaussian distribution.

\vspace{+0.2cm}
\noindent \textbf{Integrators of Diffusion Models.} To derive the generation samples from the differential equation-based diffusion models, we can calculate the integral of the reverse ODE trajectory in Eqs.~\eqref{eq:Solution} and \eqref{eq:RevODE} as $\mathbf{X}_0 = \mathbf{X}_T + \int_T^0\left[ f(\mathbf{X},t) - \frac{1}{2} g^2(t)\mathbf{S}_{\theta}(\mathbf{X}_t) \right] dt$.
Based on the semi-linear property of the diffusion models, DPM-Solver~\cite{lu2022dpm,lu2022dpm2} gives an exact solution as \vspace{-0.5em} 
\begin{equation}
    \label{Eq:DPMSolver}
    \mathbf{X}_{t-\Delta t} = \frac{\alpha_{t - \Delta t}}{\alpha_{t}}\mathbf{X}_{t} - \alpha_{t - \Delta t} \int_{\lambda_t}^{\lambda_{t - \Delta t}} e^{-\lambda}  \bm{\epsilon}_{\bm{\theta}}(\mathbf{X}_\tau,\tau) d \lambda, \vspace{-0.5em} 
\end{equation}
where $\lambda := \log( \frac{\alpha_t}{\sigma_t})$ represents the log-SNR (Signal-to-Noise Ratio); $\bm{\epsilon}_{\bm{\theta}}(\cdot)$ indicates the noise estimation neural network. By computing this integral using the Taylor series, we can finally derive the reverse diffusion results.

\vspace{+0.2cm}
\noindent \textbf{Consistency Models} are introduced to train diffusion models that can consistently estimate high-quality samples from different noise levels~\cite{consistency1,consistency2,lu2024simplifying}. Specifically, the consistency model can be parameterized as \vspace{-0.5em}  
\begin{equation}
    \mathcal{F}_{\bm{\theta}}(\mathbf{X}_t, t) = \mathbf{c}_k(t) \mathbf{X}_t + \mathbf{c}_o(t) \bm{\epsilon}_{\bm{\theta}}(\mathbf{X}_t, t), \vspace{-0.5em} 
    \label{equ:CM_para}
\end{equation}
where $\mathbf{c}_k(\Delta t)= 1$ and $\mathbf{c}_o(\Delta t)=0$ ($\Delta t >0$ and is sufficient small) to ensure the model meets the boundary condition of $\mathcal{F}_{\bm{\theta}}(\mathbf{X}_0, 0):= \mathbf{X}_0$; and $\bm{\epsilon}_{\bm{\theta}}(\cdot)$ represents a learnable neural network. Note that the ODE-Solver formulation in Eq.~\eqref{Eq:DPMSolver} can also be treated as \textit{a special case of} consistency model, only if we set $\Delta t = t$, i.e., projecting each data point back to clean state $t =0$. And we utilize this formulation in the rest of our paper. Consistency models can be trained from scratch or in a distillation manner. However, both methods aim to minimize the inconsistencies of estimations between adjacent steps, i.e., \vspace{-0.5em} 
\begin{equation}
    \label{Eq:Consistency_Distillation}
   \mathcal{L}_c =  d\left(\mathcal{F}_{\bm{\theta}}(\mathbf{X}_t, t), \mathcal{F}_{\bm{\theta}^-}(\mathbf{X}_{t-\Delta t}, t-\Delta t)\right), \vspace{-0.5em} 
\end{equation}
where $\mathbf{X}_{t-\Delta t}$ can be derived by solving inverse steps from $\mathbf{X}_t$ using a pre-trained diffusion model or be directly interpolated with the same noise sample as $\mathbf{X}_t$; $\bm{\theta}^-$ indicates to apply the stop gradient operation on the running mean of $\bm{\theta}$; and $d(\cdot)$ represents a discrepancy measurement, e.g., Frobenius Norm~\cite{bottcher2008frobenius} or LPIPS~\cite{zhang2018unreasonable}.

\begin{figure}[t]
  \centering
   \includegraphics[width=1.0\linewidth]{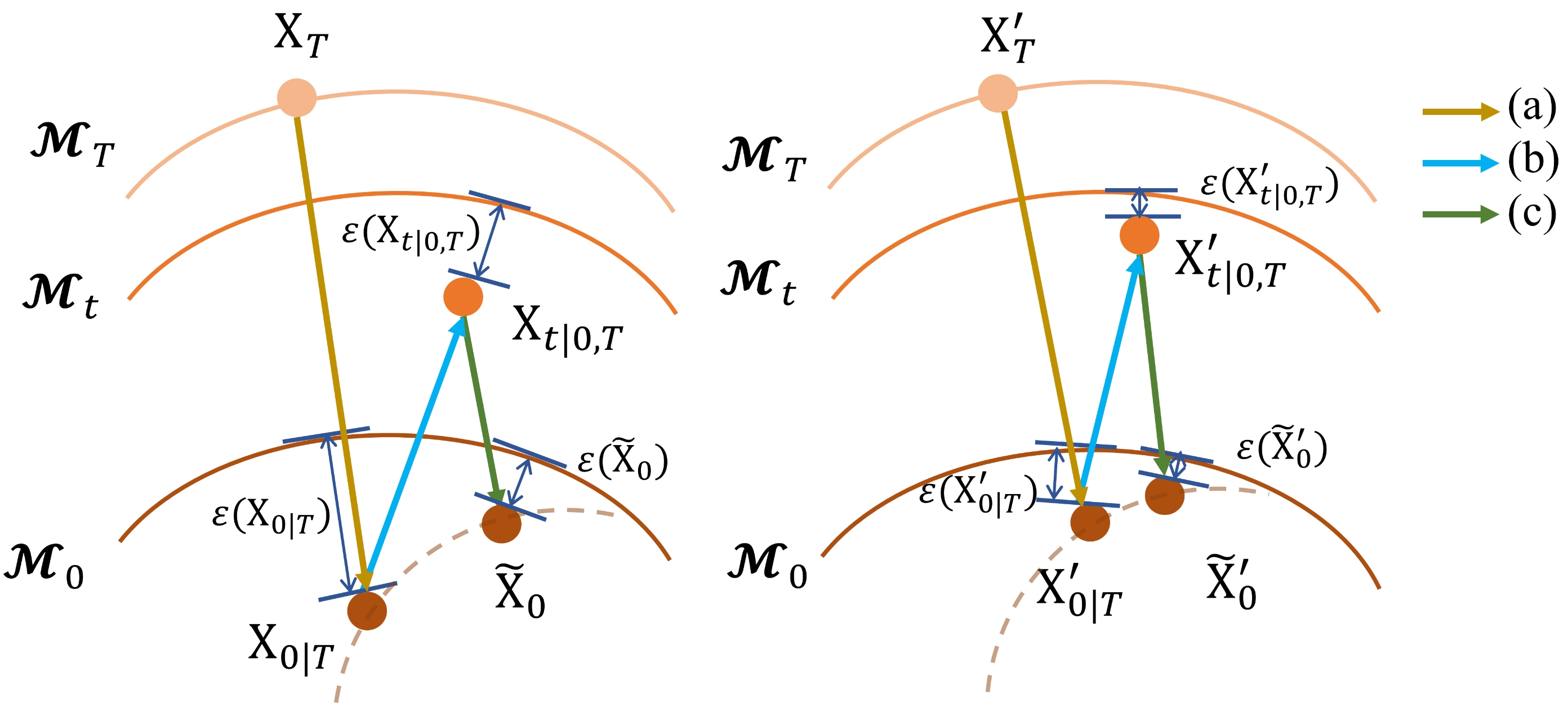}\vspace{-0.3cm}
   \caption{Illustration of the progressive score matching by our edge consistency distillation, in a data manifold view, where $\mathcal{M}_t$ indicates the data manifold with the timestamp $t$, e.g., $\mathcal{M}_T$ and $\mathcal{M}_0$ represent the manifolds of pure noise and clean samples, respectively; $\mathbf{X}_{0|T}$ represents the few-step estimation of the clean sample from Gaussian noise $\mathbf{X}_T$;
   $\widetilde{\mathbf{X}}_0$ represents a relatively accurate training target of $\mathbf{X}_{0|T}$, refined by the edge consistency region; and $\mathcal{E}(\cdot)$ is the generation error, e.g., distance between its corresponding manifold surface. 
   (a) shows the single-step reverse trajectory by the endpoint (pure noise) score function; (b) represents the noised latent interpolation using noise $\mathbf{X}_T$ and $\mathbf{X}_{0|T}$; and  (c) indicates the score estimation in the region with consistency characteristic.  
   The \textbf{left} subfigure illustrates that \underline{before} being adapted by the proposed strategy. The \textbf{right} subfigure shows that \underline{after} being trained by our method, the error of the score function at the initial pure noise state is gradually decreased as $\mathcal{E}(\mathbf{X}_{0|T})>\mathcal{E}(\mathbf{X}_{0|T}')$. It indicates that the result $\mathbf{X}_{0|T}$ can gradually approach the data manifold. 
   } 
   \label{fig:manifold}
   \vspace{-0.5cm}
\end{figure}

\section{Proposed Method}

Accurately estimating the score function is particularly challenging in states of pure random noise (i.e., the low-SNR edge). As a result, in existing diffusion-based single image-to-3D models, the few-step reverse process suffers from accumulated errors in the score function, leading to imprecise results at the final stage of denoising and significantly degrading generation performance. 
To address this issue, we propose Acc3D, which progressively improves the estimation of the score function at the endpoint (low-SNR side) during training, thus enhancing generative performance in a few-step generation.
To be specific, we propose an edge consistency model as a score corrector, enabling regularization and distillation for enhanced accuracy of score estimation at endpoints, as outlined in Sec.~\ref{Sec:EdgeConsistency}. Building on these insights, we introduce an adversarial regularization process in Sec.~\ref{Sec:Adversarial}, aimed at further minimizing the discrepancy between generated and real samples. These two modules work in synergy, strengthening each other to improve generative performance.

\vspace{0.1cm}
\subsection{Edge Consistency-guided Distillation}
\vspace{0.1cm}
\label{Sec:EdgeConsistency}
In this section, we propose edge consistency-guided distillation to reduce integral errors in the one-step inference outcomes of diffusion models. 

Technically, let $\mathbf{X}_{0|T}$ be the one-step coarse generation result from noisy latent at timestamp $T$ to $0$ through the first-order approximation of ODE-Solver in Eq.~\eqref{Eq:DPMSolver}, written as\footnote{
As discussed below in Eq.~\eqref{equ:CM_para}, the ODE-Solver in Eq.~\eqref{Eq:DPMSolver} can be thought of as a special case of consistency model. 
}
\begin{equation}
\mathbf{X}_{0|T} = \mathcal{F}_{\bm{\theta}_G}(\mathbf{X}_T,~T),
\label{equ:coarse target}
\end{equation}
where $\bm{\theta}_G$ represents the parameters of the generator. Let $\mathbf{X}_0^*$ be the on-manifold sample corresponding to $\mathbf{X}_{0|T}$, and $\mathcal{E}(\mathbf{X}_{0|T}) = \|\mathbf{X}_{0|T} - \mathbf{X}_0^* \|_F$ the error between $\mathbf{X}_{0|T}$ and $\mathbf{X}_0^*$, as depicted on the left side of Fig.~\ref{fig:manifold}.  
Effectively and accurately minimizing the value of $\mathcal{E}$ thus becomes the paramount objective to improve the quality of $\mathbf{X}_{0|T}$. 
However, obtaining $\mathbf{X}_0^*$ directly is challenging or even intractable:
(\textbf{1}) taking the clean image $\mathbf{X}_0$ as $\mathbf{X}_0^*$ causes the model to predict blurry values, approximating a weighted average of $\mathbf{X}_0^*$, since multiple $\mathbf{X}_0^*$ can be denoised by the diffusion model to produce the same noise; and
(\textbf{2}) preparing the noise-$\mathbf{X}_0^*$ pairs by generating the entire training dataset from noise is a computationally intensive process.
Thus, instead of directly deriving the sample $\mathbf{X}_0^*$, we draw inspiration from the reward lifting process in reinforcement learning and propose to regularize $\mathbf{X}_{0|T}$ to approach a feasible sample $\widetilde{\mathbf{X}}_0$ with a reduced error, i.e., $\mathcal{E}(\widetilde{\mathbf{X}}_0)\textless\mathcal{E}(\mathbf{X}_{0|T})$. With the progression of model iterations, the sample $\widetilde{\mathbf{X}}_0$ is expected to gradually improve to be closer to the data manifold surface $\mathcal{M}_0$, while the coarse estimation $\mathbf{X}_{0|T}$ steadily approaches $\widetilde{\mathbf{X}}_0$, leading to continuous refinement.  
Such a process is intuitively depicted in the right sub-figure of Fig.~\ref{fig:manifold}.

Building on this progressive refinement intuition, the focus shifts to obtaining an accurate sample $\widetilde{\mathbf{X}}_0$. 
Leveraging the consistency model's ability to generate high-quality outputs directly from the latent space with varying noise levels, we train our accelerated model with consistency constraints in Eq.~\eqref{Eq:Consistency_Distillation} within the edge region (i.e., the region with high SNR), which we term "edge consistency." The accelerated model is trained to generate the refined estimation $\widetilde{\mathbf{X}}_0$ under the guidance of edge consistency constraints. 
As $\widetilde{\mathbf{X}}_0$ experiences the additional denoising step, it becomes more accurate and closer to $\mathbf{X}^*_0$ compared to the coarse estimation $\mathbf{X}_{0|T}$. Specifically, the feasible refined estimation $\widetilde{\mathbf{X}}_0$ is computed as follows:
\begin{equation}
    \label{Eq:Distillation}
    \widetilde{\mathbf{X}}_0 = \mathcal{F}_{{\bm{\theta}_G}^{-}}(\mathbf{X}_{t|0, T}, t),\vspace{-0.3em}
\end{equation}
where $\mathbf{X}_{t|0, T}$ denotes the forward diffusion process to evolve $\mathbf{X}_{0|T}$ to timestamp $t$ using the initial noise $\mathbf{X}_T$. Here we utilize the same noise of $\mathbf{X}_T$ to keep the correspondence of data points between the noise and data manifolds. Through distilling the $\mathbf{X}_{0|T}$, we can progressively correct the coarse estimation to approach the more accurate sample $\widetilde{\mathbf{X}}_0$, i.e.,
\vspace{-0.6em}
\begin{equation}
\label{Eq:Cons_guided_distill}
    \mathcal{L}_d = d( {\widetilde{\mathbf{X}}}_0,~\mathbf{X}_{0|T}).\vspace{-0.3em}
\end{equation}
Distilling the coarse estimation with consistency guidance enhances the accuracy of the score function in the one-step accelerated model, ultimately facilitating the generation of high-quality samples.

\vspace{0.5em}
\noindent\textbf{Remark.} We are not aimed at devising novel consistency training techniques. Instead, we utilize consistency constraints to train the high-SNR edge region as an effective refinement tool to guide the learning of the score function in the pure noise state (endpoint), which can give reasonable regularization and preserve generation diversity. Our edge consistency-guided score distillation method 
strategically avoids consistency regularization in low-SNR challenging regions, potentially reducing the learning burden and thus enhancing generative performance. See the results in Table~\ref{tab:ablations} for the advantage of our edge consistency over the traditional consistency model that applies consistency regularization to the entire region. We also refer readers to Sec.~A of the \textit{Supplementary Material} for a detailed theoretical analysis.

\begin{algorithm}[t]
\caption{Training Pipeline of the Proposed Algorithm}
\label{alg1}
\begin{algorithmic}[1]  
    \Statex \textbf{Input:} Pre-trained diffusion model parameterized with $\bm{\theta}$; training dataset $\mathcal{S}$; number of iterations $K$; learning rate $\eta_G$ and $\eta_D$.
    \Statex \textbf{Output:} Optimized generation model parameters $\bm{\theta}_G^*$

    \State $\bm{\theta}_G \gets \bm{\theta}$ \Comment{Initialize $G$ from pre-trained model}
    \State Initialize discriminator $\bm{\theta}_D$ randomly.
    \For{$k = 1$ to $K$}
        \State Sample data $\mathbf{X}_0$ from $\mathcal{S}$, draw guassian noise $\bm{\epsilon} \sim \mathcal{N}(\mathbf{0}, \mathbf{I})$, draw guassian noise $\mathbf{X}_T \sim \mathcal{N}(\mathbf{0}, \mathbf{I})$
        \State Sample timestep $t \sim \mathcal{U}(0, N)$ where $N < T$ and interval $\Delta t \sim \mathcal{U}(0, \delta]$ \Comment{sample $t$ and $\Delta t$ for edge consistency}
    
        \State Interpolate noised latent: $\mathbf{X}_t = \alpha_t \mathbf{X}_0 + \sigma_t \bm{\epsilon}$ and corresponding $\mathbf{X}_{t-\Delta t} = \alpha_{t-\Delta t} \mathbf{X}_0 + \sigma_{t-\Delta t} \bm{\epsilon}$
   
      \State Compute edge consistency loss $\mathcal{L}_c$  \text{Eq.~\eqref{Eq:Consistency_Distillation}}
      \State Calculate coarse target using Eq. \eqref{equ:coarse target} 
      \State Calculate refined target $\widetilde{\mathbf{X}}_0$ using \text{Eq.~\eqref{Eq:Distillation}}
      \State Compute score distillation loss $\mathcal{L}_d$ using \text{Eq.~\eqref{Eq:Cons_guided_distill}} 
      \State Evaluate adversarial loss $\mathcal{L}_{\text{GAN}}$ using \text{Eq.~\eqref{Eq:GAN}}
      \State Update $G$: $\bm{\theta}_G \gets \bm{\theta}_G - \eta_G \nabla_{\bm{\theta}_G}(\mathcal{L}_c + \mathcal{L}_d + \mathcal{L}_{\text{GAN}})$
      \State Update $D$: $\bm{\theta}_D \gets \bm{\theta}_D + \eta_D \nabla_{\bm{\theta}_D}\mathcal{L}_{\text{GAN}}$
    \EndFor
    \State \Return $\bm{\theta}_G^*$
\end{algorithmic}
\end{algorithm}

\begin{figure*}[t]
  \centering
   \includegraphics[width=0.9\linewidth,height=0.27\linewidth]{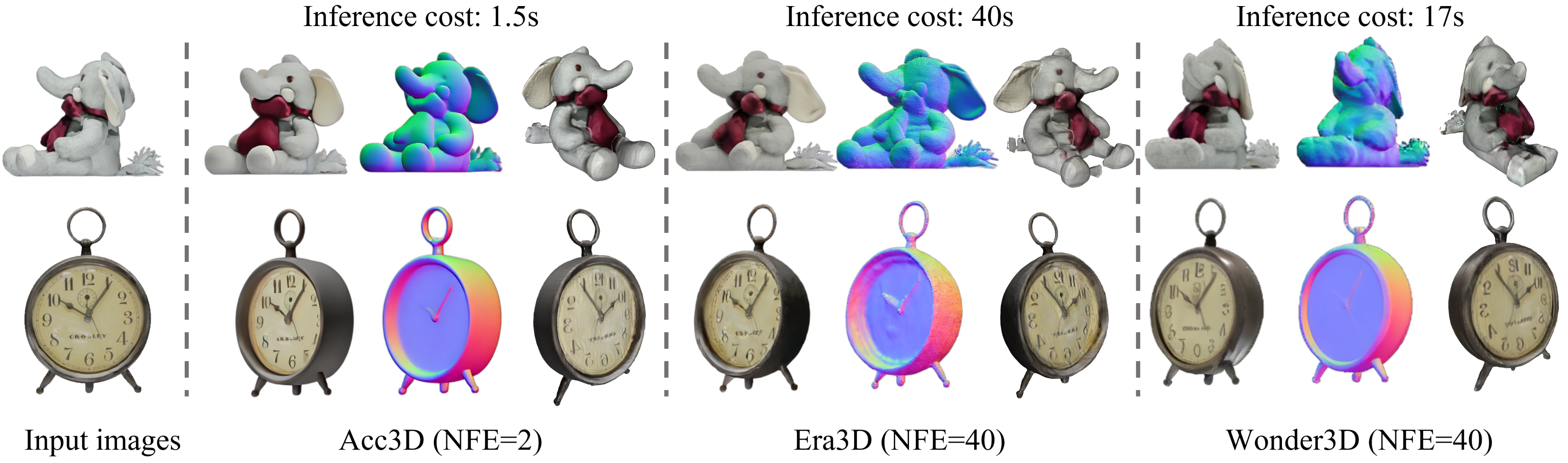}\vspace{-0.1cm}
   \caption{Visual comparisons of our Acc3D, Era3D \cite{era3d}, and Wonder3D \cite{wonder3d} on the GSO~\cite{gso} dataset.
   For each sample, we provide the generated view, normal map, and reconstructed 3D mesh, displayed from left to right, respectively.}
   \vspace{-0.3cm}
   \label{fig:gso}
\end{figure*}
\vspace{0.1cm}
\subsection{Disentangled Adversarial Regularization}
\label{Sec:Adversarial}
To further calibrate the distribution of the generated data, in this section, we introduce the adversarial training technique to directly measure the error of generation by a learnable discriminator~\cite{sauer2025adversarial}. As shown in Fig.~\ref{fig:pipeline}, to leverage the pre-trained geometric knowledge in the diffusion model, we utilize the pre-trained diffusion model as the discriminator. However, due to the huge distribution gap between the geometric (normal) and texture components, in this process, we separate the discriminative learning processes via introducing a dual-discriminator, as shown in Fig.~\ref{fig:pipeline}. Moreover, the adversarial learning objective can be formulated as
\begin{equation}
    \label{Eq:GAN}
    \mathcal{L}_{GAN} = \mathcal{D}_{\bm{\theta}_D}(\mathcal{F}_{\bm{\theta}_G}(\mathbf{X}_T, T)) - \mathcal{D}_{\bm{\theta}_D}(\mathbf{X}^*),
\end{equation}
where $\bm{\theta}_{\mathcal{G}}$ and $\bm{\theta}_{\mathcal{D}}$ are the network parameters of the generator and discriminator, respectively, $\mathcal{F}_{\bm{\theta}_G}(\mathbf{X}_T, T) = \mathbf{X}_{0|T}$ and $\mathbf{X}^*$ indicates the real images and normal.

\vspace{0.5em}
\noindent \textbf{Remark.} Decoupling the learning of geometry and texture information is a simple yet effective strategy that enhances the adversarial training process. We also want to highlight that while leveraging adversarial training can significantly enhance generative capabilities, without robust guidance—such as our edge consistency-guided progressive distillation detailed in Sec. \ref{Sec:EdgeConsistency}—adversarial learning is prone to instability and mode collapse, as demonstrated in the ablation studies outlined in Table~\ref{tab:ablations} and Fig.~\ref{fig:ablations}. In essence, the decoupled adversarial regularization and edge consistency complement each other, mutually reinforcing to elevate generative performance.

In all, Algorithm~\ref{alg1} summarizes the training pipeline of the proposed method.

\section{Experiments}
\subsection{Experimental Settings}
\noindent \textbf{Dataset.}
Our model was trained on the Objaverse~\cite{objaverse} dataset. We have aligned our training dataset with that of Wonder3D~\cite{wonder3d}, which includes approximately 35,000 objects.
We prepared our rendered multiview images and normal maps with Blender protocols.
The rendered results consist of six views, all adjusted to the same scale as Era3D, including the front, back, left, right, front-right, and front-left views. For the evaluation dataset, we adopt the Google Scanned Object dataset, following prior works~\cite{syncdreamer, wonder3d}. For the quantitative evaluation, we initially rendered the input image at a resolution of $512\times512$, subsequently obtaining multiview results and 3D assets. 

\vspace{0.45em}
\noindent \textbf{Evaluation Metrics.} We evaluated our Acc3D from multiple perspectives, including the quality of novel view synthesis and the consistency of 3D reconstruction. Following ~\citet{syncdreamer}, we used common metrics for image quality assessment to evaluate the performance of generated multiview outputs, such as PSNR, LPIPS, and SSIM. Additionally, we utilize more unpaired and perceptual scores, e.g., MUSIQ, CLIPIQA and MANIQA, to validate the performance of our model.
For single-view 3D reconstruction, we reported Chamfer Distances (CD) and Volume IoU between reconstructed shapes and ground-truth shapes. This demonstrated the rationality and multiview consistency of the image-to-3D diffusion model.

\begin{table}[t]
\renewcommand{\arraystretch}{1.0}
\setlength{\tabcolsep}{2pt} 
\centering
\small
\caption{\label{tab:nvs} Quantitative comparison of different methods for single view reconstruction. The best and second-best results are highlighted in \textbf{bold} and \underline{underlined}, respectively. 
}
\vspace{-0.12cm}
\resizebox{.45\textwidth}{!}{
\begin{tabular}{M{0.8cm}M{0.8cm}rM{0.8cm}M{0.9cm}M{0.8cm}|M{0.8cm}M{0.8cm}}
\toprule
 \multicolumn{1}{c|}{} & Ze.123 & Sync. & Wo.3D & Era3D & One23.   & \textbf{Acc3D} & \textbf{Acc3D}  \\ 
 \multicolumn{1}{c|}{Metrics}  & \cite{zero123} & ~\cite{syncdreamer} & \cite{wonder3d} & \cite{era3d} & \cite{One-2-3-45}    & \textbf{Ours} &\textbf{Ours}  \\ \toprule
     \multicolumn{1}{c|}{ \textcolor{gray}{NFE $\downarrow$}}    & 50  & 50 & \underline{40}  & \underline{40} & 50 & \textbf{2} & \textbf{4}\\ 
     \hline
\multicolumn{1}{c|}{} & \multicolumn{5}{c|}{\textbf{\textcolor{gray}{GSO Dataset}}~\cite{gso}} \\ 
\hline
\multicolumn{1}{c|}{ \textcolor{gray}{MUSIQ $\uparrow$}}     &60.39  &64.46  &58.19  &63.64  &61.79    &\underline{67.85} &\textbf{69.26}  \\
\multicolumn{1}{c|}{ \textcolor{gray}{CLIPIQA $\uparrow$}}     &0.48  &0.52  &0.45  &0.53  &0.43    & \underline{0.62} &\textbf{0.65}   \\ 
\multicolumn{1}{c|}{ \textcolor{gray}{MANIQA  $\uparrow$}}     &0.56  &\underline{0.63}  &0.55 &0.42 &0.54    & 0.59 &  \textbf{0.64}  \\ 
\hline
\multicolumn{1}{c|}{} & \multicolumn{5}{c|}{\textbf{\textcolor{gray}{DTC Dataset}}~\cite{dtc}} &\multicolumn{1}{c}{}\\
\hline
\multicolumn{1}{c|}{ \textcolor{gray}{PSNR $\uparrow$}}   &17.84  & 22.31 &23.02  &\underline{23.36}  & 16.01      & 23.06  & \textbf{24.01} \\
\multicolumn{1}{c|}{ \textcolor{gray}{SSIM $\uparrow$}}  &0.78  & 0.24 & 0.86   &\underline{0.87} &0.75    & \underline{0.87} &\textbf{0.88}  \\
\multicolumn{1}{c|}{ \textcolor{gray}{LPIPS $\downarrow$}}   & 0.25  & \textbf{0.10} & 0.14 &0.14  &0.35       &0.14 & \underline{0.12} \\
\multicolumn{1}{c|}{ \textcolor{gray}{MUSIQ $\uparrow$}}    & 58.15 &56.67  &55.24  &60.13  &59.39  &\underline{64.46} & \textbf{65.52} \\
\multicolumn{1}{c|}{ \textcolor{gray}{CLIPIQA $\uparrow$}}     &0.51  &0.55  &0.47  &0.56  & 0.46   & \underline{0.63}  &\textbf{0.65}\\ 
\multicolumn{1}{c|}{ \textcolor{gray}{MANIQA $\uparrow$}}     &0.56  &\textbf{0.64}  &0.54  &0.42  &0.53  & 0.56   &\underline{0.59} \\ 
\bottomrule
\end{tabular}
}
\vspace{-0.4cm}
\end{table}

\begin{figure*}[t]
  \centering
\includegraphics[width=0.85\linewidth,height=0.55\linewidth]{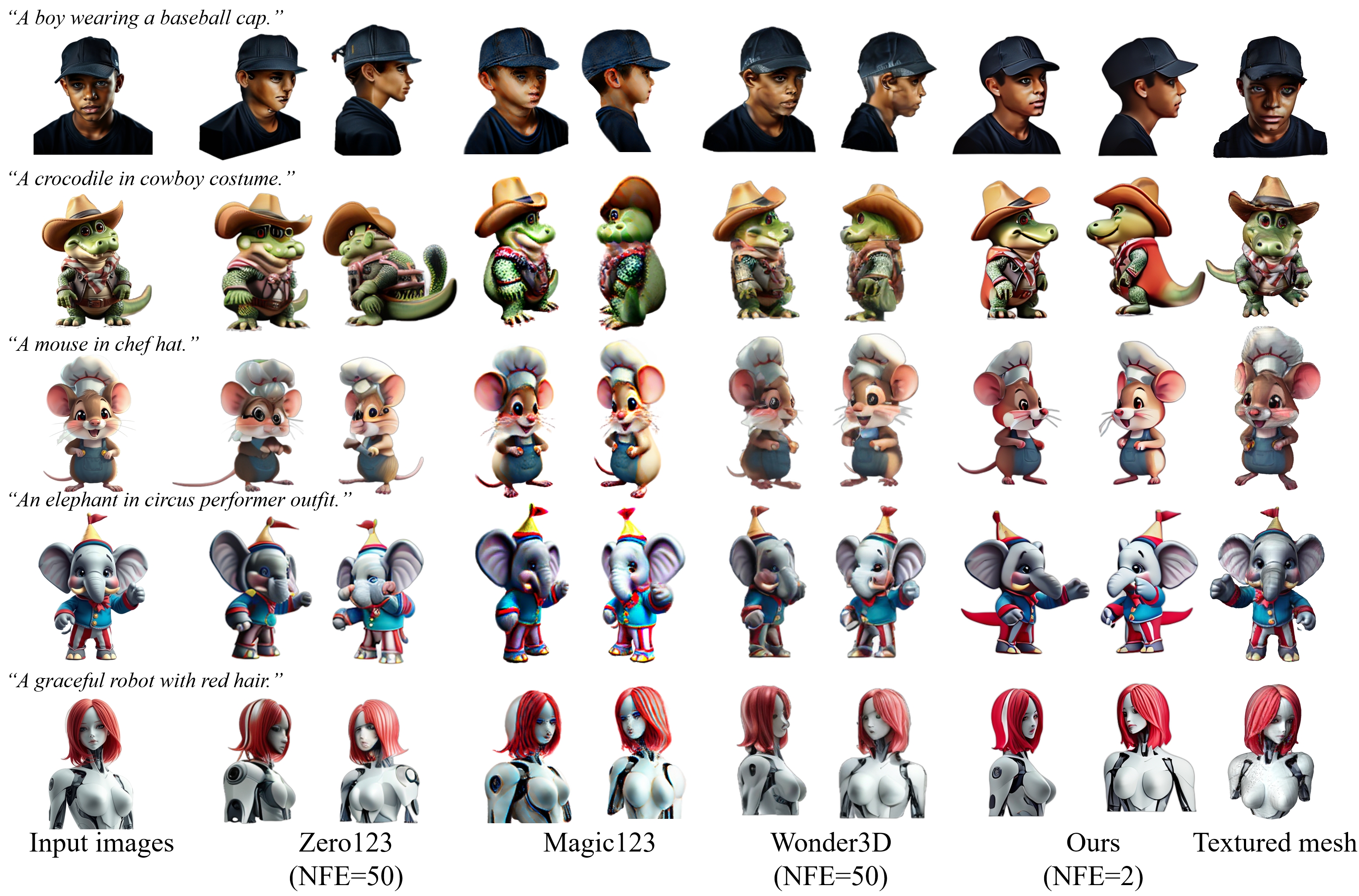}\vspace{-0.3cm}
   \caption{\label{fig_comparisons}The qualitative results generated by Acc3D on various styles of images generated by text-to-image model Flux~\cite{flux}. Please refer to Fig.~\ref{fig:vsbase} for more comprehensive comparisons between our Acc3D and baseline model Era3D. \color{blue}{\faSearch~} Zoom in for details.} \vspace{-0.3cm}
\end{figure*}

\vspace{0.4em}
\noindent \textbf{Baselines.}
We compared our Acc3D with 11 state-of-the-art image-to-3D models, including Zero123~\cite{zero123}, RealFusion~\cite{sds4}, Magic123~\cite{magic123}, One-2-3-45~\cite{One-2-3-45}, Point-E~\cite{point-e}, Shap-E~\cite{shap-e}, SyncDreamer~\cite{syncdreamer}, Wonder3D~\cite{wonder3d}, Era3D~\cite{era3d}, as well as two accelerated image-to-3D models, Stable-Fast-3D~\cite{sf3d} and TripoSR~\cite{triposr}. RealFusion and Magic123 use SDS loss on Stable Diffusion for single-view reconstruction. Zero123, SyncDreamer, Wonder3D, and Era3D generate consistent multiview images, with Wonder3D and Era3D also producing normal maps.

\vspace{0.4em}
\noindent \textbf{Implementation Details.}
We adapt the proposed model on the recent effective 3D generation method Era3D~\cite{era3d}, which can concurrently generate normal maps and multiview images at a high resolution of $512\times512$. We used AdamW as the optimizer with an initial learning rate set to $5e-6$, and the number of gradient accumulation steps of 8. The training process generally necessitates 24 hours on a cluster equipped with four 40GB NVIDIA GeForce RTX A6000 GPUs. For 3D reconstruction, we followed Era3D to generate 3D assets using NeuS~\cite{wang2021neus}.

\begin{figure}[t]
  \centering
   \includegraphics[width=0.92\linewidth]{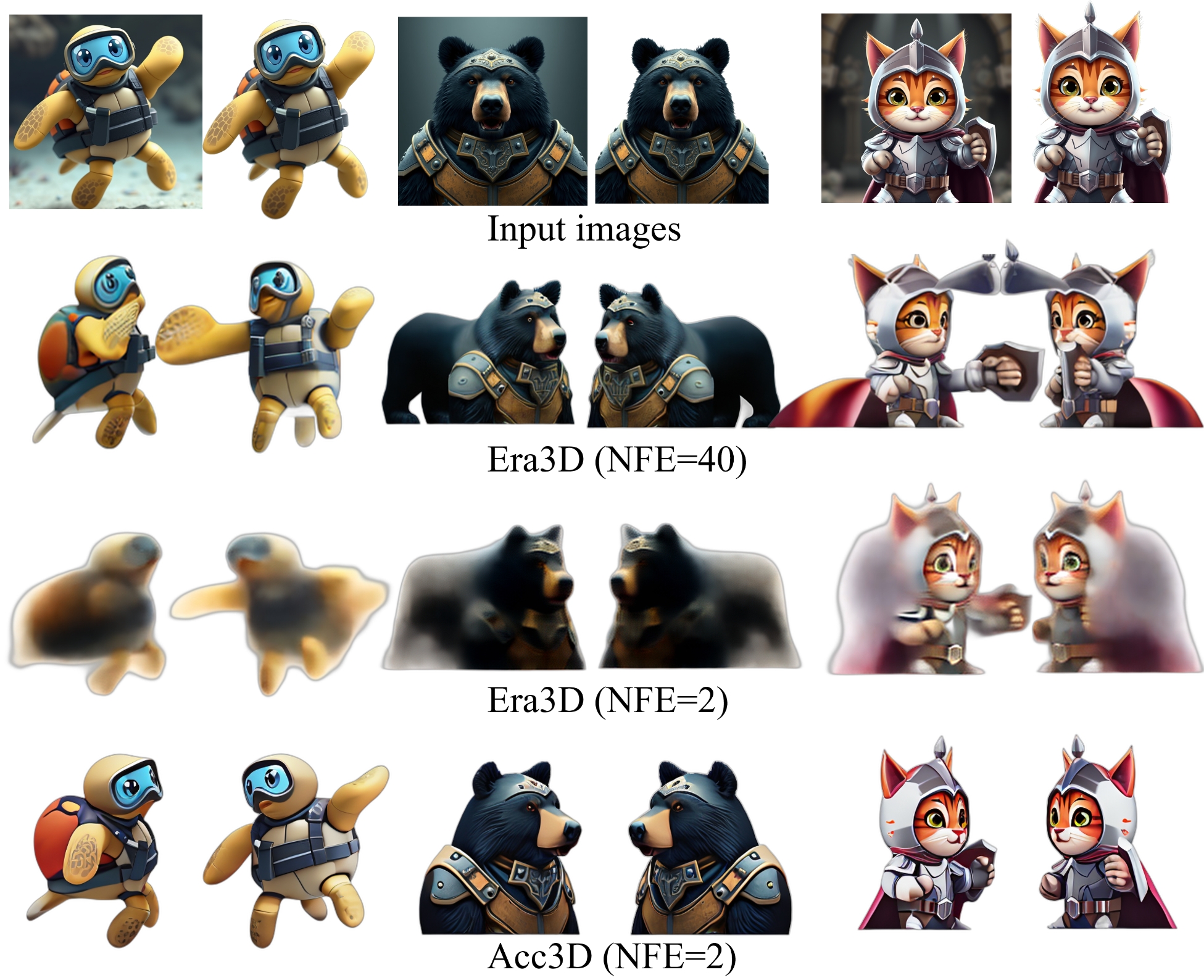}\vspace{-0.1cm}
   \caption{Visual results of our Acc3D and Era3D (baseline model). 
   }
   \label{fig:vsbase}
   \vspace{-0.5cm}
\end{figure}

\begin{table}[]
\renewcommand{\arraystretch}{1.12}
\setlength{\tabcolsep}{8pt} 
\centering
\caption{\label{tab:cd_iou} Quantitative results for 3D reconstruction on GSO. Symbol $^\dagger$ indicates the method that directly generates 3D meshes.}
\vspace{-0.15cm}
\resizebox{.4\textwidth}{!}{
\begin{tabular}{c|M{1.2cm}M{1.0cm}c}
\toprule
{Metrics} & \textcolor{gray}{\textbf{CD} $\downarrow$} & \textcolor{gray}{\textbf{IoU} $\uparrow$} & \textcolor{gray}{\textbf{NFE} $\downarrow$} \\ \hline
Realfusion~\cite{sds4} & 0.0819 & 0.2741 & -- \\ 
One-2-3-45~\cite{One-2-3-45}& 0.0629 & 0.4086 & 50\\ 
Point-E~\cite{point-e} & 0.0426 &0.2875   & 64 \\ 
Shap-E~\cite{shap-e} & 0.0436 &0.3584   & 64  \\ 
Magic123~\cite{magic123} & 0.0516 &0.4528   & -- \\ 
Zero123~\cite{zero123} & 0.0339 &0.5035   &  50 \\ 
SyncDreamer~\cite{syncdreamer} &0.0261 &0.5421  & 50   \\ 
Wonder3D~\cite{wonder3d} &0.0199 &0.6244  & 40  \\ 
Era3D~\cite{era3d} &0.0217 &0.5973  & 40  \\ 
InstantMesh~\cite{instantmesh} & 0.0406 &0.4778 & 75\\ 
Stable-Fast-3D$^\dagger$~\cite{sf3d} & 0.0421&  0.5141 & -- \\ 
TripoSR$^\dagger$~\cite{triposr} &0.0326 & 0.5326 & -- \\
\hline
\textbf{Acc3D (Ours)} &\textbf{0.0191}  &\textbf{0.6681}  & \textbf{2} \\ 
\bottomrule
\end{tabular} 
}
\vspace{-0.48cm}
\end{table}

\subsection{Experimental Results}
\noindent \textbf{Novel View Synthesis Results.} We present the numerical results in Table~\ref{tab:nvs} to quantitatively evaluate the quality of generated novel views. Our model is assessed on both the GSO~\cite{gso} and DTC~\cite{dtc} datasets, with the DTC dataset being the most recent collection that offers a high degree of granularity and intricate 3D details. 
Acc3D achieves the highest average scores on most metrics, especially in natural image quality, highlighting our framework's superiority in novel view synthesis.
We further visually compare different examples from the GSO dataset in Fig.~\ref{fig:gso}. Our method achieves superior visualization results while reducing inference time. 
Besides those aforementioned datasets with reference multiviews, we also evaluate our method in the wild using T2I-generated images in Fig.~\ref{fig_comparisons}.
Our method consistently generates high-definition, detailed results compared to other methods, especially on the challenging samples of the crocodile ($2^{nd}$ row) and the face of the elephant ($4^{th}$ row). 
We also compare our method with base model Era3D in Fig.~\ref{fig:vsbase}. Our accelerated model achieves better results with NFE of 2, while Era3D struggles with few steps.

\vspace{0.46em}
\noindent \textbf{3D Reconstruction Results.}
We conduct both quantitative and qualitative evaluations of the reconstructed geometry as presented in Table~\ref{tab:cd_iou} and Fig.~\ref{fig:gso}. In Table~\ref{tab:cd_iou}, our method attains excellent reconstruction results with the least number of function evaluations (NFE). 
Fig.~\ref{fig:gso} shows that Wonder3D is highly sensitive to the facing direction of input images; for instance, it fails to generate novel views for the sample of elephant in the left pose. Despite boasting a higher resolution of 512, Era3D still produces relatively coarse mesh outputs.

\begin{table}[]
\renewcommand{\arraystretch}{1.0}
\setlength{\tabcolsep}{0.62pt} 
\centering
\caption{\label{tab:ablations} Quantitative results of ablation studies on the GSO dataset. $\checkmark$ (resp. $\times$) indicates the presence (resp. absence) of the corresponding component. $\uparrow$ (resp. $\downarrow$) means the larger (resp. the smaller), the better. ``\textit{Distill.}" stands for score distillation with Eq.~\eqref{Eq:Distillation}. "\textit{Consis.}" represents the timestamp region of consistency loss, corresponding to $[0,~N]$ in Alg.~\ref{alg1}. $[0, T]$ indicates the consistency regularization is applied to all noise levels during training. ``\textit{Adv.}" refers to adversarial learning. ``Single" indicates the utilization of only one discriminator for both normal and texture components. The metrics were obtained under \textbf{two-step} inference.
}
\resizebox{.48\textwidth}{!}{
\begin{tabular}{M{1.15cm}M{0.8cm}cc|cM{0.8cm}cM{1.1cm}}
\toprule
& \textit{Distill.} & \textit{Consis.} & \textit{Adv.}  & \textcolor{gray}{\textbf{PSNR}$\uparrow$}  & \textcolor{gray}{\textbf{CD}$\downarrow$} & \textcolor{gray}{\textbf{MUSI.}$\uparrow$} & \textcolor{gray}{\textbf{CLIPI.}$\uparrow$} \\ \hline \textit{(a)}& $\times$ & [0,~$T$] & $\times$  &18.11  &0.060  & 61.01 &0.55 \\
\textit{(b)}& $\times$ & [0,~$T$] & $\checkmark$  &20.79  &0.021  &67.22  & \textbf{0.62}\\
\textit{(c)}& $\checkmark$ & [0,~$T$] & $\checkmark$  &21.10  &0.025  &65.39  & 0.60\\
\textit{(d)}& $\checkmark$ & $\times$ & $\checkmark$  &20.47  &0.026  &63.08 &0.58 \\ 
\textit{(e)} & $\times$ & [0,~0.4$T$] & $\checkmark$ & 20.32 &0.027  &66.57  &\textbf{0.62} \\
\textit{(f)}& $\checkmark$ & [0,~0.2$T$] & $\checkmark$  &21.22 &0.027  &67.24  &0.58   \\
\textit{(g)}& $\checkmark$ & [0,~0.6$T$] & $\checkmark$  &22.13  &0.022  & 66.87  &0.59 \\ 
\textit{(h)}& $\checkmark$ & [0,~0.8$T$] & $\checkmark$  &21.23  &0.024  &65.06  &0.61 \\ 
\textit{(i)}&$\checkmark$ & [0,~0.4$T$] & $\times$  &21.09  & 0.029 &51.05  &0.45 \\ 
\textit{(j)}& $\checkmark$ &[0,~0.4$T$] & single  & 21.37 & 0.022 & 66.41 & 0.57 \\  \hline
Era3D\cite{era3d}& -- & -- & -- & 20.17  &0.023  &53.81  &0.46  \\
\textbf{Ours}  & $\checkmark$ & [0,~0.4$T$] & $\checkmark$ & \textbf{22.47}  & \textbf{0.019}  & \textbf{67.85} & \textbf{0.62}\\
\bottomrule
\end{tabular}
}
\vspace{-0.3cm}
\end{table}

\subsection{Ablation Studies}

We perform comprehensive ablation studies (Table~\ref{tab:ablations}), training all models similarly except for the ablation term and evaluating on GSO dataset.

\begin{figure}[t]
  \centering
   \includegraphics[width=0.9\linewidth,height=0.7\linewidth]{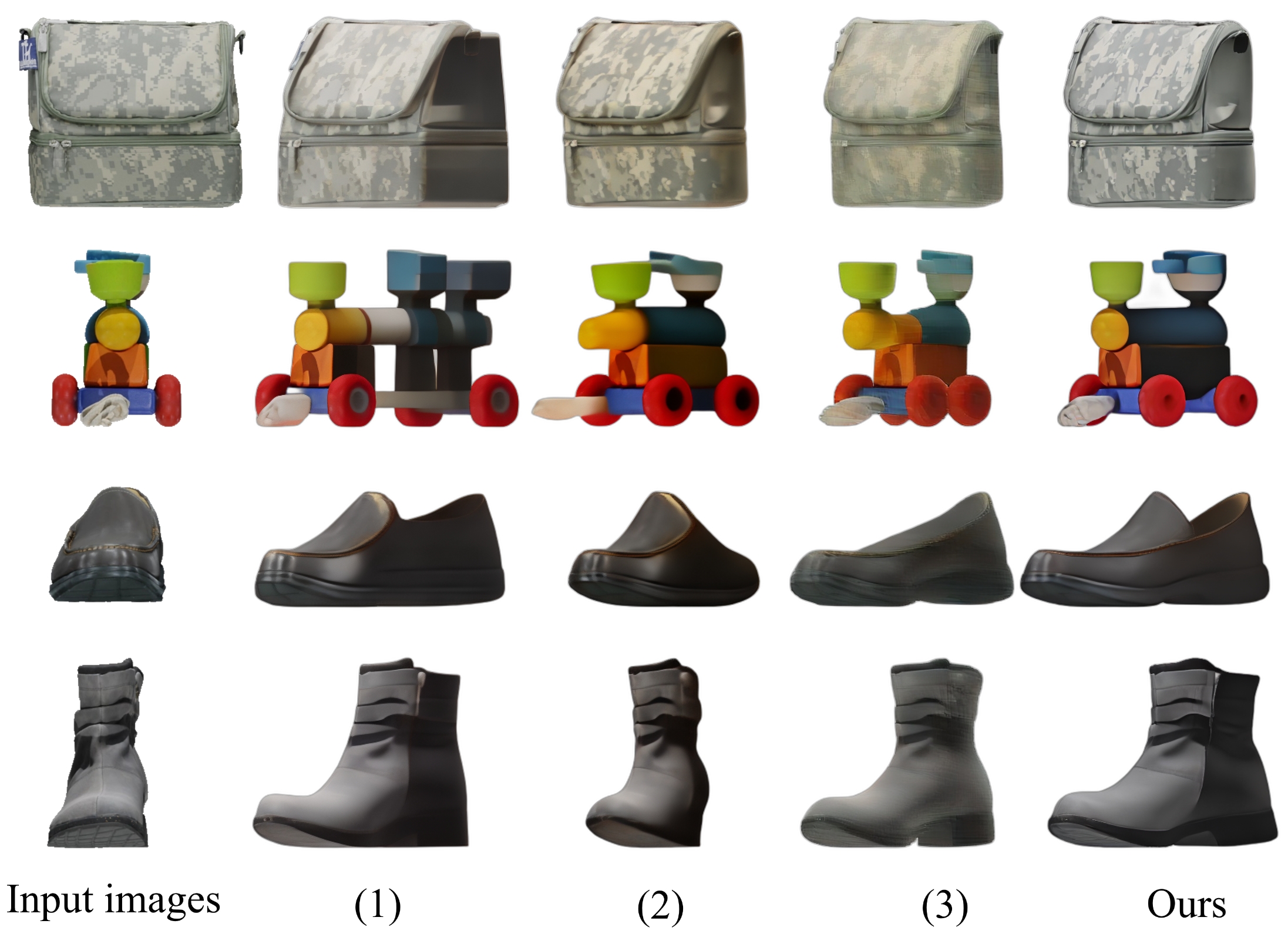}\vspace{-0.1cm}
   \caption{Visualizations of the ablation study conducted on GSO. All results are generated in two steps. The results (1), (2), and (3) correspond to the experimental configurations (b), (g), and (i) outlined in Table~\ref{tab:ablations}, respectively.}
   \label{fig:ablations}
   \vspace{-0.5cm}
\end{figure}

\vspace{0.46em}
\noindent \textbf{Disentangled Adversarial Learning.}  
Through comparing Table~\ref{tab:ablations} (j) and ``Ours", we find that separately learning different modalities can help stabilize adversarial learning and get better quantitative results.
As depicted in Fig.~\ref{fig:ablations} (3) and ``Ours", the fusing of these two modalities adversely affect the discriminator/model's performance, shown obviously for the $1^{st}$ and $3^{rd}$ samples.

\vspace{0.46em}
\noindent \textbf{Various Distillation Settings.} 
We evaluate the effectiveness of our consistency-guided score distillation. Specifically, comparisons between Table~\ref{tab:ablations} (e) and ``Ours" validate the \textit{necessity of distillation}. Without distillation loss, being solely trained with adversarial learning is hard to derive high-quality results. Table~\ref{tab:ablations} (a)/(b) \textit{V.S.} ``Ours" also indicates the superiority of our edge consistency-guided distillation compared with full consistency model~\cite{consistency1,consistency2}. We can also visually compared with Fig.~\ref{fig:ablations} (1) and ``Ours" that the setting of (b) generates strange samples, especially for the bag in the $1^{st}$ row. This suggests that GANs are prone to mode collapse without robust regularization.

\vspace{0.46em}
\noindent \textbf{Parameter selections of consistency constraints.} By comparing Table~\ref{tab:ablations} (c), (d), (f), (g), (h) and ``Ours", 
we can conclude that using a consistency model to guide score matching at the endpoint is more effective than directly utilizing the original diffusion model. Additionally, the choice of the region for the consistency constraint is crucial. Selecting an appropriate edge consistency region can significantly enhance the generation results

\section{Conclusion}

We introduced Acc3D, a novel and streamlined method designed to accelerate single image-to-3D diffusion models. Acc3D integrates a consistency-guided distillation process and an adversarial augmentation strategy. Our model significantly improves computational efficiency, achieving a speed increase of over $20 \times$, while also enhancing the quality of the generated 3D models. Our method outperforms the baseline and most other image-to-3D models in metrics, such as CD and IoU. 
The effectiveness and efficiency of our proposed approach highlight its potential for broad adoption. All the code and datasets utilized in our research are publicly available, fostering further investigation and refinement by the broader scientific community.

\clearpage
{
    \small
    \bibliographystyle{ieeenat_fullname}
    \bibliography{main}
}





\setcounter{section}{0}
\renewcommand\thesection{\Alph{section}}

\maketitle
\thispagestyle{empty}  
\renewcommand{\theequation}{S.\arabic{equation}}

\vspace{-5em}
\section{Theoratical Investigation of Edge Consistency V.S. Full Consistency Distillation}
As mentioned in the manuscript, the consistency model regularizes the consistency of estimated clean samples as 
\begin{equation}
    \mathcal{L}_c = \left\| \mathcal{F}_{\bm{\theta}}(\mathbf{X}_t, t) - \mathcal{F}_{\bm{\theta}^-}(\mathbf{X}_{t-\Delta t}, t-\Delta t) \right\|_F.
\end{equation}
\noindent The original consistency models hypothesize that when the consistency training loss $\mathcal{L}_c$ decreases to $0$, the error of consistency function follows $\mathcal{O}((\Delta t)^p)$. However, there is a critical issue that such hypnosis is too ideal for large-scale model and dataset training processes. The original loss $\mathcal{L}_c$ cannot be optimized to $0$. Thus, there is an inevitable accumulated error, shifting the estimations of the consistency model from the original data manifold.  

To further take such optimization error into consideration, we assume that the error, e.g., the difference between two step inference results, follows a uniform distribution, i.e., $\mathbf{u} = \mathcal{F}_{\bm{\theta}}(\mathbf{X}_t, t) - \mathcal{F}_{\bm{\theta}^-}(\mathbf{X}_{t-\Delta t}, t-\Delta t)$, and $ \mathbf{u}  \sim \mathcal{U}\left(0,\delta\right)$, where $\delta$ is a quite small number.

\vspace{0.3em}
\noindent \textbf{Notions}. We denote the consistency function of empirical PF ODE as $\mathcal{F}_{\bm{\Phi}}(\cdot, \cdot)$ with $\bm{\Phi}$ as the parameters of well-trained (ideal) diffusion model, which can be realized by the integral process. We then calculate the error term $\left\| \mathcal{F}_{\bm{\theta}}(\bm{x}_{t_n}, t_n) - \mathcal{F}_{\bm{\Phi}}(\bm{x}_{t_n}, t_n)\right\|_F$. We denote $\mathcal{E}_n$ as the error term with the timestamp $t_n$.

\begin{equation}
    \mathcal{E}_n := \mathcal{F}_{\bm{\theta}}(\bm{x}_{t_n}, t_n) - \mathcal{F}_{\bm{\Phi}}(\bm{x}_{t_n}, t_n).
\end{equation}
Moreover, for the term with timestamp $t_{n+1}$ , we have 
\begin{align}
    \mathcal{E}_{n+1} &= \mathcal{F}_{\bm{\theta}} (\bm{x}_{t_{n+1}}, t_{n+1}) - \mathcal{F}_{\bm{\Phi}}(\bm{x}_{t_{n+1}}, t_{n+1}); \nonumber \\
    &=\mathcal{F}_{\bm{\theta}}(\hat{\mathbf{x}}_{t_n}^{\bm{\phi}}, t_n) + \bm{u} - \mathcal{F}_{\bm{\Phi}}(\mathbf{x}_{t_{n}}, t_{n}); \nonumber \\
    &=\mathcal{F}_{\bm{\theta}}(\hat{\mathbf{x}}_{t_n}^{\bm{\phi}}, t_n) -\mathcal{F}_{\bm{\theta}}(\mathbf{x}_{t_n}, t_n) + \bm{u}_n \\
    & \quad + \mathcal{F}_{\bm{\theta}}(\mathbf{x}_{t_n}, t_n) - \mathcal{F}_{\bm{\Phi}}(\bm{x}_{t_{n}}, t_{n}); \nonumber \\
    &=\mathcal{F}_{\bm{\theta}}(\hat{\mathbf{x}}_{t_n}^{\bm{\phi}}, t_n) -\mathcal{F}_{\bm{\theta}}(\mathbf{x}_{t_n}, t_n) + \bm{u}_n + \mathcal{E}_n. \nonumber
\end{align}
Then, we have 
\begin{align}
    & \| \mathcal{E}_n \|_F \leq \nonumber \\
    & \| \mathcal{E}_1 \|_F + \sum_{k=1}^{n} \left(\|\bm{u}_k\|_F + \| \mathcal{F}_{\bm{\theta}}(\hat{\mathbf{x}}_{t_k}^{\bm{\phi}}, t_k) -\mathcal{F}_{\bm{\theta}}(\mathbf{x}_{t_k}, t_k)\|_F\right);  \nonumber\\
    & \overset{\textcircled{1}}{=} n\times \mathbb{E}_{\mathbf{x}_{t_n}\sim\mathbf{P}(\mathbf{x}_{t_n})} \left[\|\bm{u}_n\|_F\right] + \sum_{k=1}^n\mathcal{O}\left( (t_{k+1} - t_k)^{p+1} \right);  \nonumber\\
    &\overset{\textcircled{2}}{\leq} n\times \mathbb{E}_{\mathbf{x}_{t_n}\sim\mathbf{P}(\mathbf{x}_{t_n})} \left[\|\bm{u}_n\|_F\right] + \sum_{k=1}^n\mathcal{O}\left( (\Delta t)^{p+1} \right); \nonumber\\
    &= n \times \left\{\mathbb{E}_{\mathbf{x}_{t_n}\sim\mathbf{P}(\mathbf{x}_{t_n})} \left[\|\bm{u}_n\|_F\right] + \mathcal{O}((\Delta t)^p)\right\} , 
\end{align}
where $\textcircled{1}$ is for the importance sampling of $\bm{u}$, $\| \mathcal{E}_1 \|_F = 0$ (the boundary condition) and Taylor expansion of $\mathcal{F}_{\bm{\theta}}$; in $\textcircled{2}$, $\Delta t = \max (t_{n+1} - t_n)$ is selected as the largest time interval. From the results, we can derive that instead of the inherent Tylor high order residues $\mathcal{O}((\Delta t)^p)$, there is further training residual error term of $N\times \mathbb{E} \|\bm{u}\|_F$. Moreover, both of them are scaled by the number of intervals $N$, which indicates the accuracy of consistency is negatively associated with the length of the interval we want to regularize, i.e., the full consistency resulting in the largest potential error.
Here we denote the aforementioned upper error boundary $\mathbb{E}_{\mathbf{x}_{t_n}\sim \mathbf{P}(\mathbf{x_{t_n}})} \| \bm{u}_n \|_F + \mathcal{O}((\Delta t)^p)$ by $\left\|\mathcal{E}_r\right\|_F$. 
We then have $\|\mathcal{E}_n \|_F \leq n \times \| \mathcal{E}_r \|_F$. 
\noindent Moreover, considering the distillation process, we feed the one-step generation result into the edge consistency region via interpolating the latent at timestamp $t$ as
\begin{equation}
    \mathbf{x}_{t|0,T} = \alpha_t \mathbf{x}_{0|T} + \sigma_t \bm{\epsilon}. \nonumber
\end{equation}
Since such one-step generation is coarse due to the error of the score function in the pure noise state, note that such interpolation is a contraction mapping for our estimation term $\mathbf{x}_{0|T}$ ($\alpha_t \leq 1$). By assuming the consistency function has $L-$Lipschitz continuity, we have
\begin{align}
    \|\mathcal{F}_{\bm{\theta}}(\mathbf{x}_t^*,t) - \mathcal{F}_{\bm{\theta}}(\mathbf{x}_{t|0,T},t) \|_F &\leq 
    L\|\mathbf{x}_t^* - \mathbf{x}_{t|0,T}\|_F, \nonumber \\
    &=L\alpha_t \|\mathbf{x}_0^* - \mathbf{x}_{t|0,T}\|_F,\nonumber
\end{align}
which indicates the error of our distillation target is positively associated with data ratio $\alpha_t$. However, note that such errors are estimated over the trained consistency model parameterized by $\bm{\theta}$, whose error is shown as $\|\mathcal{E}_n \|_F$. Thus, taking both consistency model training error with inherent noised latent interpolation error into consideration, we have
\begin{align}
    \mathcal{E}_A (\mathbf{x}_{0|t}) &\leq L\alpha_t \|\mathbf{x}_0^* - \mathbf{x}_{t|0,T}\|_F + n \times \| \mathcal{E}_r \|_F, \nonumber\\
    &\overset{\textcircled{1}}{=} L\alpha_t \|\mathbf{x}_0^* - \mathbf{x}_{t|0,T}\|_F + t \times \| \mathcal{E}_r \|_F, \nonumber 
\end{align}
where $\textcircled{1}$ is derived by setting the distillation step $t$ to be the same as the largest edge consistency training timestamp $n$. Then, it's natural to derive a lower error upper bound as setting the $t^*$ to ensure $\frac{d \alpha_t}{dt}|_{t = t^*} = -\frac{L\alpha_t \|\mathbf{x}_0^* - \mathbf{x}_{t|0, T}\|_F }{ \| \mathcal{E}_r \|_F}$. It indicates that we need to train and utilize the consistency in the region of $[ t^*, 0 ]$, which is exactly the proposed edge consistency. Moreover, we utilize empirical experiments to validate the choice of $t^*$, as shown in our ablation studies.

\section{Details of the Adversarial Architecture}
We extract the features from the last three layers of the diffusion model. The feature channels are processed to attain a uniform channel value of 640, achieved through the use of a $1\times 1$ convolution. Next, average pooling is used to ensure consistent sizing across the features. The three layers of features are then concatenated along the channel dimension and fed into the prediction head. Inspired by DMD~\cite{dmd2}, the prediction head is composed of a series of 4x4 convolutions with a stride of 2, group normalization, and SiLU activations. All feature maps are downscaled to a $4\times 4$ resolution, which is then followed by a singular convolutional layer with a kernel size and stride of 4. This layer aggregates the feature maps into a single vector, which is subsequently fed into a linear projection layer to predict the classification score.

\begin{figure}[!htb]
  \centering
\includegraphics[width=0.9\linewidth]{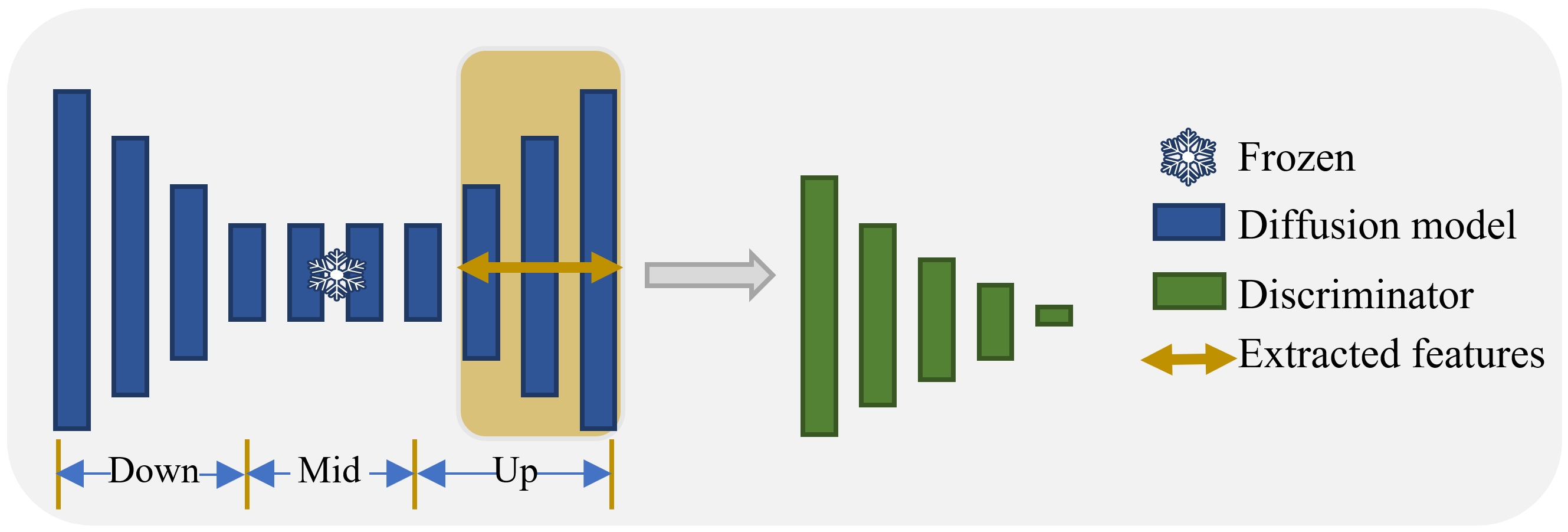}
   \caption{\label{fig:supp_discriminator} Illustration of the adversarial architecture, where we utilize the features from the last three layers of the feature extractor (the features in the yellow box).}  
\end{figure}

\section{Implementation Details}
The training phase of the Generative Adversarial Network (GAN) is unstable and susceptible to influence from the initial phase. Hence, we warm up the single-step diffusion model using a typical distillation framework to distill knowledge from the multi-step diffusion model to the single-step one, to avoid the failure of single-step prediction. Specifically, the multi-step model executes inference in 6 steps and the warm-up phase trains approximately 6,000 iterations.

\section{More Visual Results}
\begin{figure*}[t]
  \centering
\includegraphics[width=1.0\linewidth,height=1.1\linewidth]{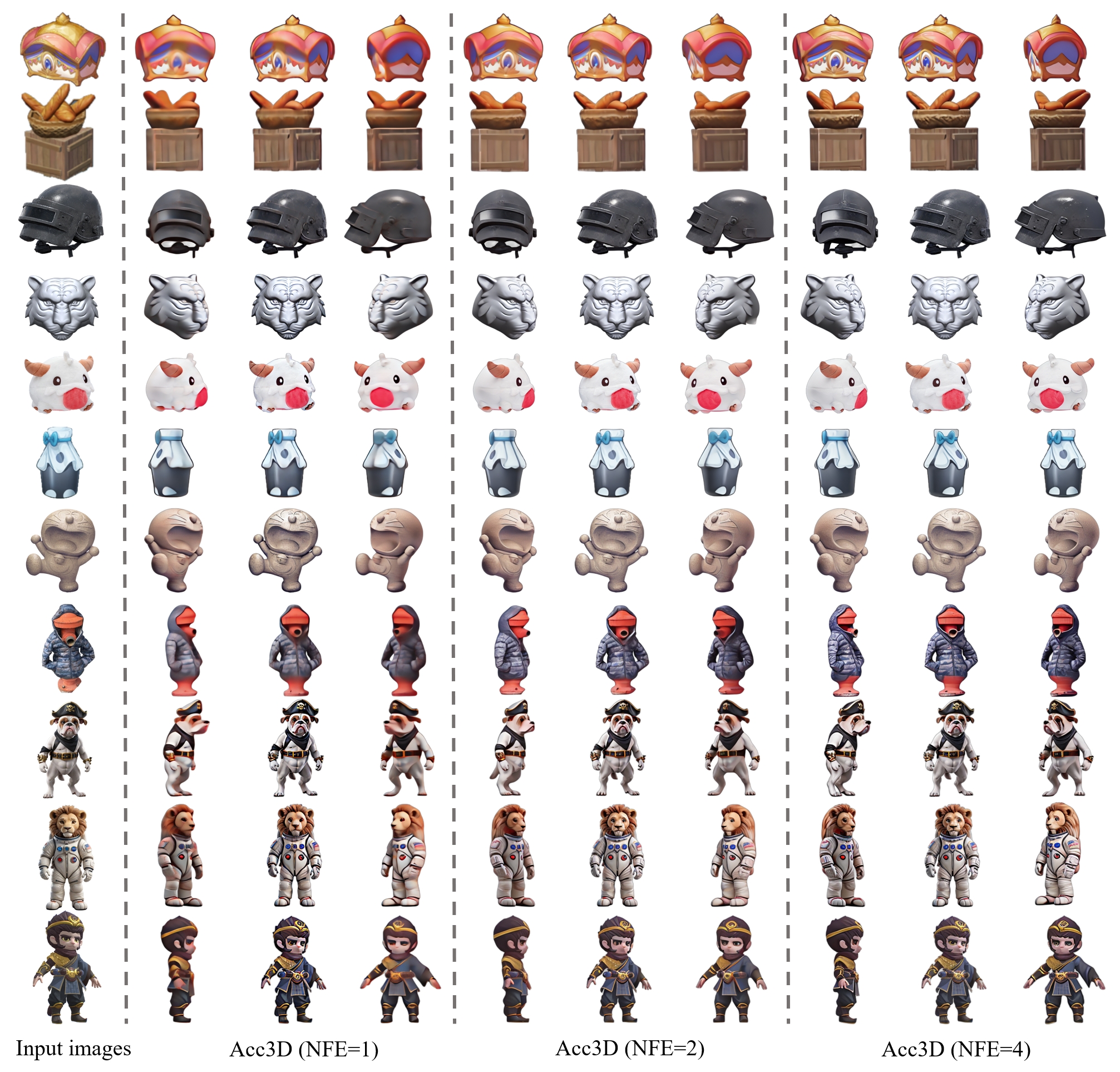}
   \caption{\label{fig:supp_more_visual}The qualitative results generated by Acc3D on typical demonstrations in Image-to-3D. Acc3D is capable of producing outstanding multi-view outputs in just two steps. \color{blue}{\faSearch~} Zoom in for details. } 
\end{figure*}

\begin{figure*}[t]
  \centering
\includegraphics[width=0.95\linewidth,height=1.1\linewidth]{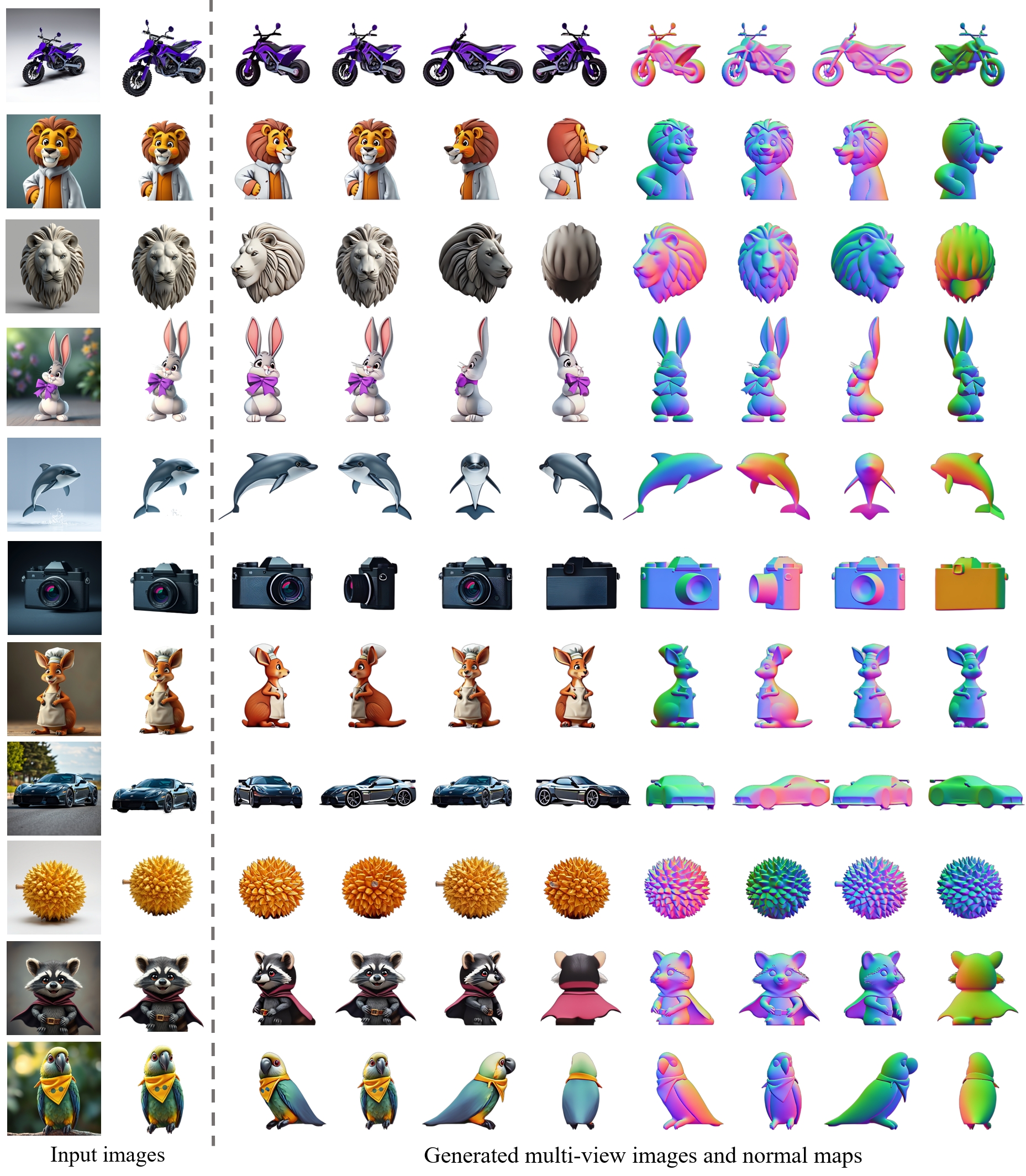}
   \caption{\label{fig:supp_more_visual2}The qualitative results produced by Acc3D, utilizing \textbf{less than four} inference steps on a variety of image styles generated by the Text-to-Image model Flux~\cite{flux}. \color{blue}{\faSearch~} Zoom in for details. }  
\end{figure*}

\begin{figure*}[t]
    \centering
    \begin{minipage}{0.475\linewidth}
        \centering
        \includegraphics[width=1.0\linewidth]{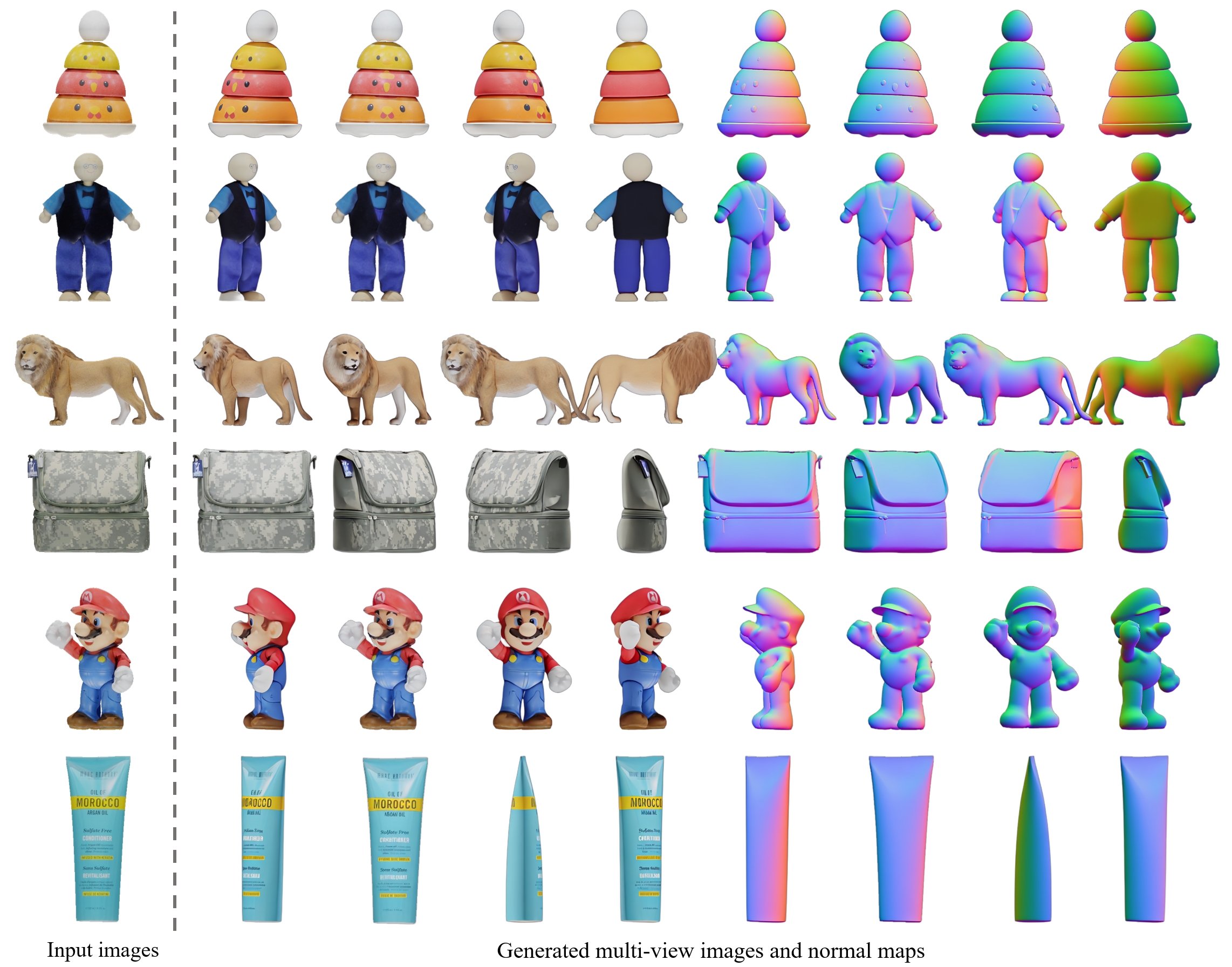}
   \caption{\label{fig:supp_gso}The qualitative results on GSO dataset~\cite{gso}. \color{blue}{\faSearch~} Zoom in for details. }  
    \end{minipage}
    \hspace{5mm}
    \begin{minipage}{0.485\linewidth}
        \centering
       \includegraphics[width=1.0\linewidth]{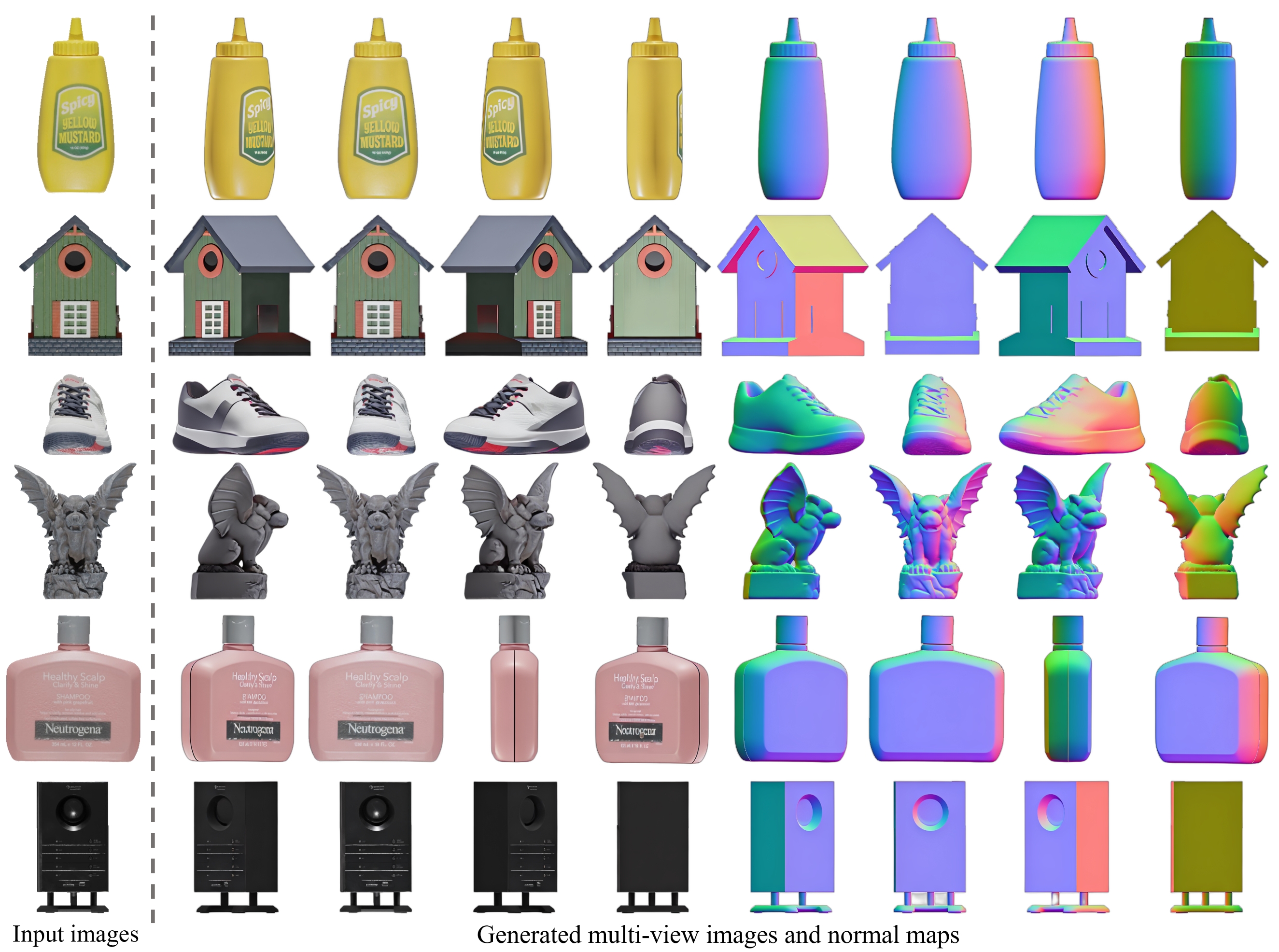}
   \caption{\label{fig:supp_dtc}The qualitative results on DTC dataset~\cite{dtc}. \color{blue}{\faSearch~} Zoom in for details. } 
    \end{minipage}
\end{figure*}

In this section, We present additional visual results generated by our accelerated model to further highlight its capabilities and performance. These examples include a range of typical demonstrations for Image-to-3D tasks, which are commonly used benchmarks in multi-view image diffusion models~\cite{era3d, wonder3d, syncdreamer}. The results, shown in Figure~\ref{fig:supp_more_visual}, demonstrate the model’s ability to produce high-quality, multiview-consistent 3D reconstructions with remarkable efficiency.
Additionally, we present further outputs produced by the Text-to-Image (T2I) model, Flux~\cite{flux}, which are visualized in Figure~\ref{fig:supp_more_visual2}. 
In the main paper, we conducted a comprehensive evaluation of our accelerated model on the GSO~\cite{gso} and DTC~\cite{dtc} datasets, which are well-established benchmarks for assessing 3D generation tasks. To provide deeper insights and reinforce the superior performance of our approach, we include additional visual results in Figure~\ref{fig:supp_gso} and Figure~\ref{fig:supp_dtc}. These results further demonstrate the model's ability to achieve high-quality, multiview-consistent 3D reconstructions with significantly fewer inference steps, setting a new benchmark for diffusion-based 3D generation methods.

\section{Comparison with Base Model Era3D} 

\begin{figure}[!htb]
  \centering
\includegraphics[width=1.0\linewidth]{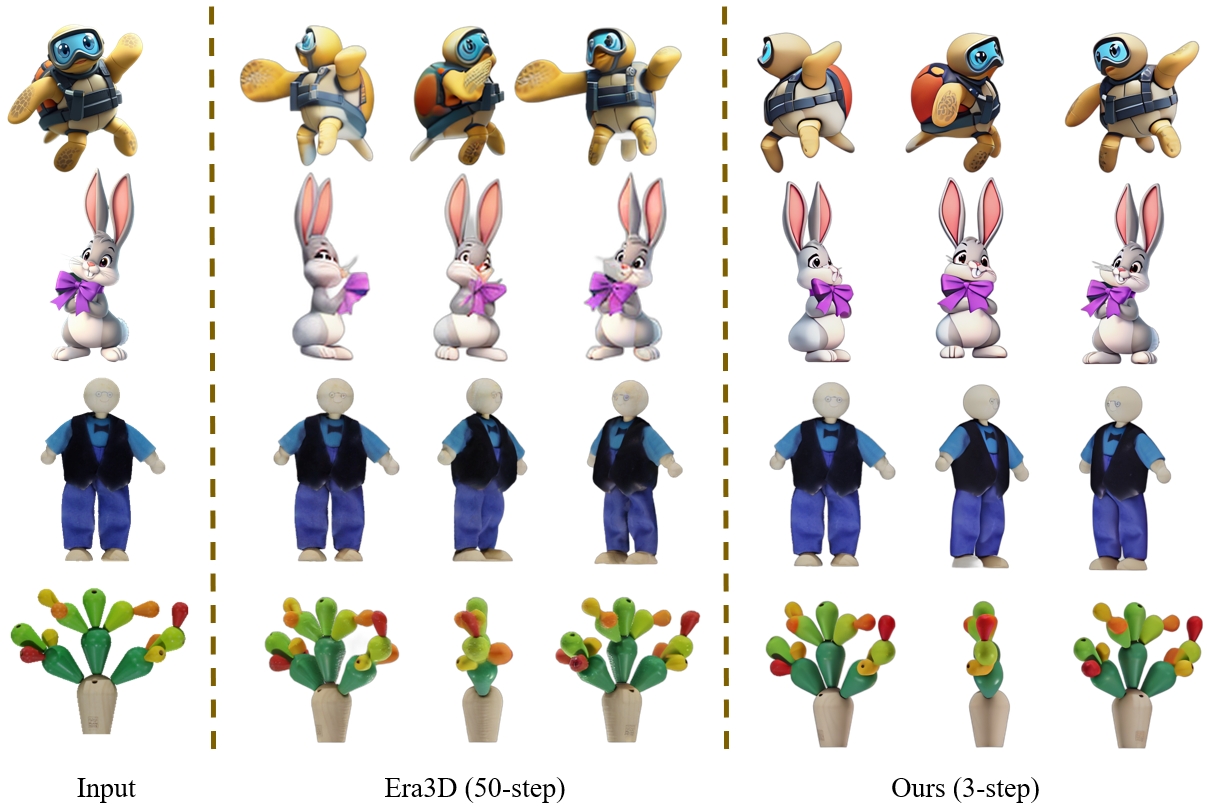}
   \caption{\label{fig:comparison_era3d} Visual comparisons with baseline model Era3D.}
\end{figure}

We visualize the results of our method and the base model Era3D in Fig.~\ref{fig:comparison_era3d}. Our method produces clearer and visually superior results at the same resolution (512), demonstrating finer details and better structural consistency. Additionally, our approach requires significantly fewer steps to achieve high-quality novel view synthesis, making it more efficient while maintaining superior visual fidelity. This highlights the effectiveness of our framework in generating high-resolution, high-quality 3D-aware images with reduced computational cost.

\begin{figure*}[t]
    \centering
    \begin{minipage}{0.475\linewidth}
        \centering
       \includegraphics[width=1.0\linewidth,height=0.5\linewidth]{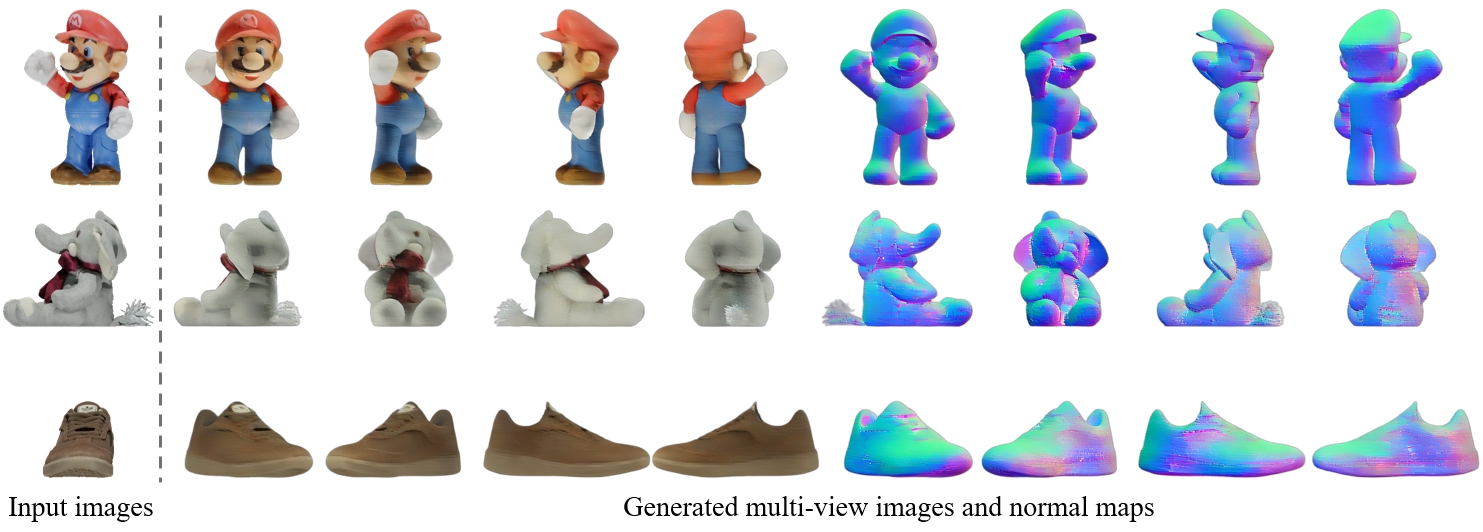}
   \caption{\label{fig:supp_single} Visualizations of single discriminator processing both colors and normal maps simultaneously.} \vspace{-0.5cm}
    \end{minipage}
    \hspace{5mm}
    \begin{minipage}{0.485\linewidth}
      \includegraphics[width=0.95\linewidth]{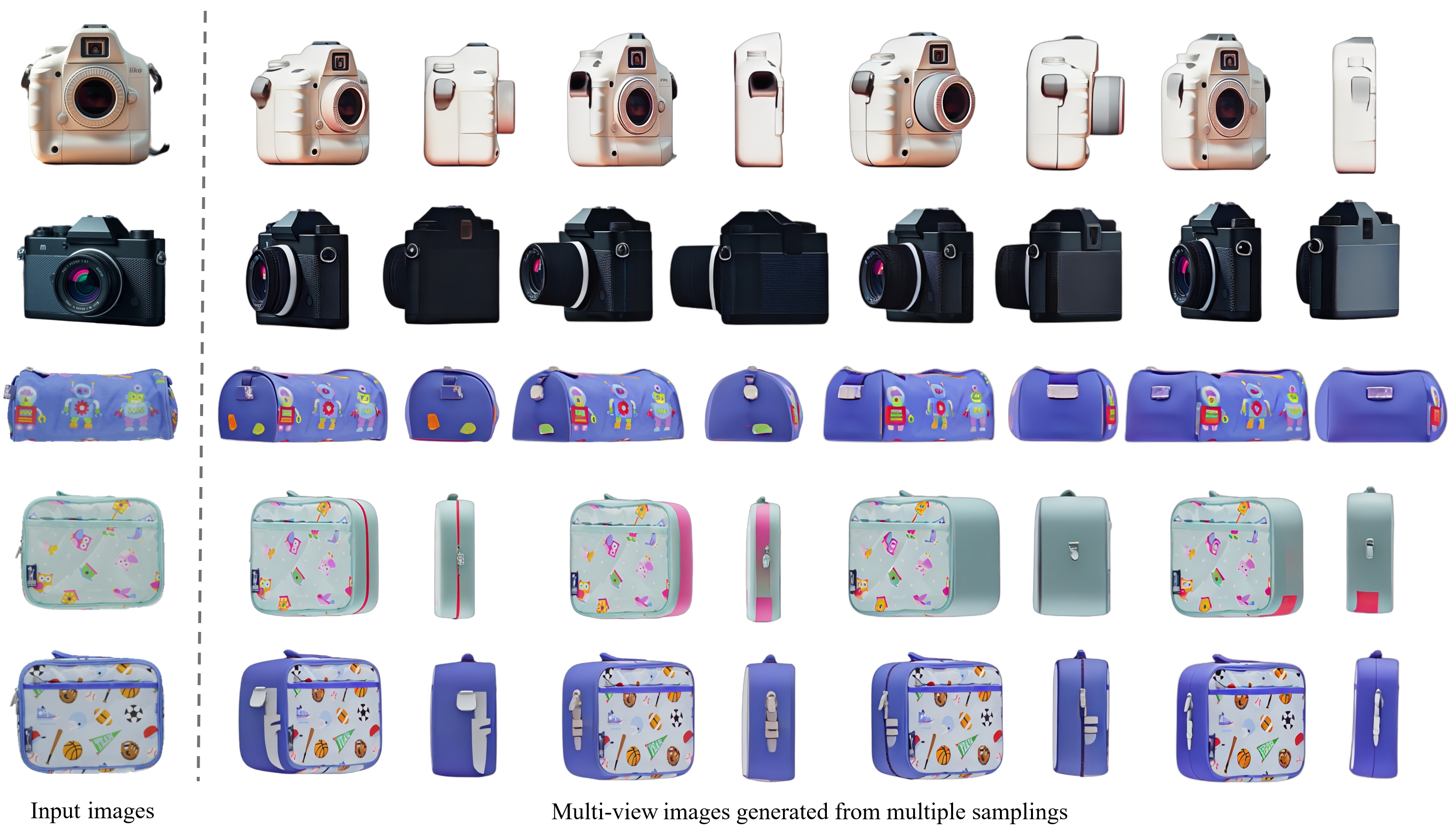}\vspace{-0.1cm}
   \caption{\label{fig:supp_multi_sampling} Diversity in the synthesis of novel views under different seeds. The diverse results highlight a broad range of diversity, capturing both the geometrical and visual characteristics that are not present in the input view.}  
    \end{minipage}
\end{figure*}

\begin{figure*}[!htb]
  \centering
\includegraphics[width=1.0\linewidth]{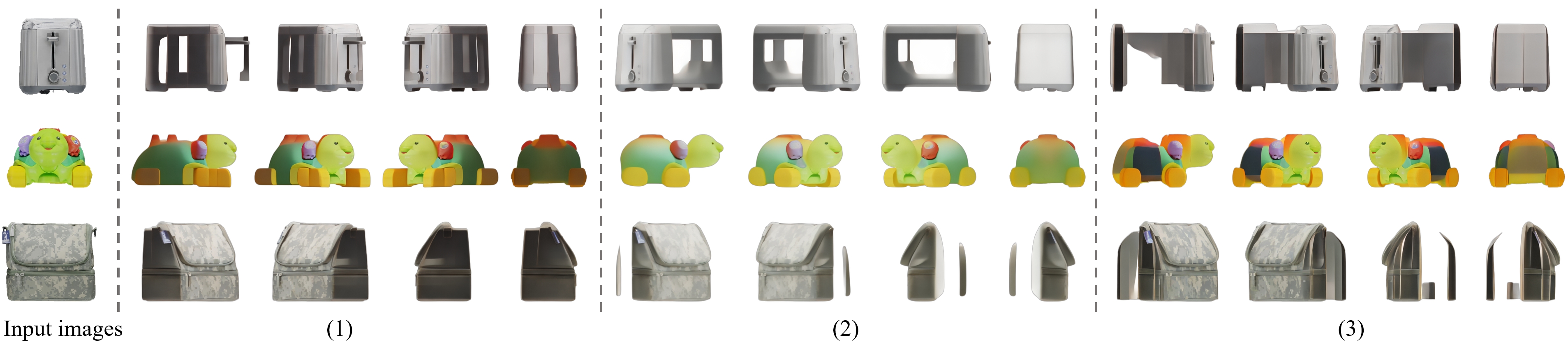}
   \caption{\label{fig:supp_collapse} Visualizations of some collapse under various experiment settings. All results are generated in two steps. The results (1), (2), and (3) correspond to the experimental configurations (b), (c), and (e) outlined in Table 3 from the main paper, respectively.}  
\end{figure*}

\section{More Visual Results of Ablation Studies}
In this section, we provide a detailed analysis of Table 3 from the main paper and present some intuitive visualization results.
\subsection{Risk of Mode Collapse}
As shown in Fig.~\ref{fig:supp_collapse} (1) and (3), in the absence of distillation, adversarial learning is prone to mode collapse, with all results skewing towards an unusual pattern. This is particularly evident with the $2^{nd}$ and $3^{rd}$ samples in (1).
Fig.~\ref{fig:supp_collapse} (2) showcases the outcomes of fully guided score distillation. As shown in $1^{st}$ and $3^{rd}$ samples in (2), fully guided distillation potentially amplifies the learning burden, subsequently lowering generative performance.

\subsection{Negative Effect of Single Discriminator}
As depicted in Fig.~\ref{fig:supp_single}, the fusing of these two modalities adversely affect the discriminator/model's performance, shown obviously for the normal maps of  $2^{nd}$ and $3^{rd}$ samples. Independent learning of different modalities in the discriminator can enhance the stability of adversarial learning and result in superior quantitative outcomes.

\section{Diversity of Sampling} 
Diversity is an important evaluation indicator for single image-to-3D model. Our model demonstrates the capability to produce an array of superior-quality examples, as depicted in Fig~\ref{fig:supp_multi_sampling}.

\end{document}